\documentclass[letterpaper]{article} % DO NOT CHANGE THIS
\usepackage{aaai2026}  % Camera-ready mode: author information is displayed
\nocopyright  % arXiv preprint: suppress publisher copyright footer
\usepackage{times}  % DO NOT CHANGE THIS
\usepackage{helvet}  % DO NOT CHANGE THIS
\usepackage{courier}  % DO NOT CHANGE THIS
\usepackage[hyphens]{url}  % DO NOT CHANGE THIS
\usepackage{graphicx} % DO NOT CHANGE THIS
\usepackage{natbib}  % DO NOT CHANGE THIS AND DO NOT ADD ANY OPTIONS TO IT
\usepackage{caption} % DO NOT CHANGE THIS AND DO NOT ADD ANY OPTIONS TO IT
\usepackage{booktabs}
\usepackage{tabularx}
\usepackage{array}
\usepackage{ragged2e}
\usepackage{xcolor}
\usepackage{fvextra}
\usepackage{algorithm}
\usepackage{algorithmic}
\usepackage{amsmath}
\newcolumntype{P}[1]{>{\RaggedRight\arraybackslash}p{#1}}
\newcolumntype{Y}{>{\RaggedRight\arraybackslash}X}

\DefineVerbatimEnvironment{PromptBlock}{Verbatim}{
  fontsize=\scriptsize,
  breaklines=true,
  breakanywhere=true,
  frame=single,
  framesep=2mm,
  rulecolor=\color{black!30},
  tabsize=2
}

\usepackage{amssymb}
\newcommand{\smallcite}[1]{{\scriptsize\citep{#1}}}
\usepackage{threeparttable}
\usepackage{newfloat}
\usepackage{listings}
\usepackage{tikz}
\DeclareCaptionStyle{ruled}{labelfont=normalfont,labelsep=colon,strut=off} % DO NOT CHANGE THIS
\floatstyle{ruled}
\newfloat{listing}{tb}{lst}{}
\floatname{listing}{Listing}
\title{D-VLC: Decentralized Vision-Language Collaboration for Heterogeneous Embodied Multi-Robot Systems in Unknown Environments}
\author{
Yuan Zhou\textsuperscript{\textdagger}\textsuperscript{\rm 1,2},
Ruitong Lin\textsuperscript{\textdagger}\textsuperscript{\rm 1,2},
Shen Wang\textsuperscript{\textdagger}\textsuperscript{\rm 1,2},
Weiqi Gai\textsuperscript{\rm 2,3},\\
Mo Zhu\textsuperscript{\rm 1,2},
Xin Zhou\textsuperscript{\rm 2},
Yuze Wu\textsuperscript{\rm *,2},
Fei Gao\textsuperscript{\rm *,1,2}
}

\affiliations{
\textsuperscript{\rm 1}Zhejiang University, Hangzhou, China\\
\textsuperscript{\rm 2}Differential Robotics, Hangzhou, China\\
\textsuperscript{\rm 3}Beihang University, Beijing, China\\
\textsuperscript{\textdagger}These authors contributed equally to this work.\\
\textsuperscript{*}Corresponding authors. 
E-mail: wuyuze000@zju.edu.cn (Yuze Wu), 
fgaoaa@zju.edu.cn (Fei Gao)
}
\begin{document}

\maketitle

\begin{abstract}
Multi-robot systems, particularly heterogeneous robot swarms, can improve the efficiency of complex task execution through parallel collaboration and complementary capabilities. However, conventional rule-based methods rely on predefined task models and specialized decision making programs, making it difficult to understand complex semantic instructions and coordinate heterogeneous robots. LLMs introduce strong language understanding and task reasoning capabilities, allowing multi-robot systems to interpret instructions, decompose tasks, and assign roles according to task semantics. VLMs further incorporate visual perception, enabling robots to reason about objects, regions, and spatial relationships in physical environments. Nevertheless, existing LLM/VLM based methods often depend on known maps, centralized and synchronized decision making, limiting their generalization to heterogeneous robots and unseen tasks. We therefore propose a framework that combines decentralized asynchronous reasoning, lightweight information sharing, capability aware collaboration, and a unified action interface, enabling general purpose VLMs to generate robot specific actions executed by learning free experts without task or robot specific training. Experiments across diverse scenarios and multiple VLMs show success rates above 70\%, with completion time reduced by up to 55.8\% relative to the geometric greedy baseline.

\end{abstract}
\section{Introduction}

\begin{figure*}[t]
    \centering
    \includegraphics[width=6.4in]{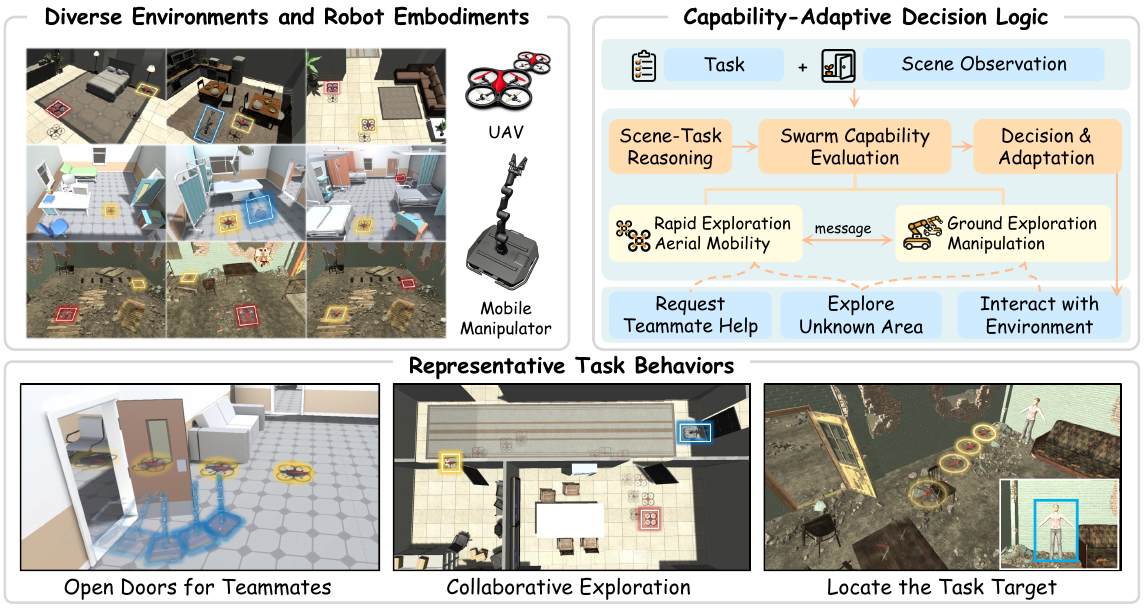}
    \caption{Decentralized Heterogeneous Multi-Robot Collaboration in Unseen Environments.}
    \label{fig1}
\end{figure*}

Multi-robot systems can improve task execution efficiency through parallel collaboration \cite{swarm_all, f_swarm, new_swarm}. In heterogeneous multi-robot systems, different types of robots can further leverage their respective strengths through capability complementarity, enabling them to accomplish complex tasks that are difficult for a single robot \cite{h_swarm}. However, existing multi-robot systems are mostly developed for specific tasks, such as exploration \cite{zby_ex}, target tracking \cite{y_track}, and collaborative mapping \cite{map_swarm}. They rely on manually predefined task models and separately designed decision-making programs for heterogeneous robots, making them difficult to adapt to diverse task requirements in open environments. In real-world applications, humans often expect to specify tasks directly through natural language, yet such instructions usually contain complex semantics and implicit constraints that are difficult to formalize in advance. Consequently, rule-based multi-robot systems remain limited in understanding task intent, coordinating heterogeneous capabilities, and generalizing to complex tasks.

To overcome the dependence of conventional rule-based methods on predefined task models and manually encoded rules, recent studies have introduced LLMs and VLMs into multi-robot systems \cite{roco, mcoconav, viki, AGrobot, icra}. With their capabilities in language understanding \cite{gai_nav}, task reasoning \cite{vln_wu}, and knowledge transfer, LLMs can decompose tasks, assign roles, and plan collaboration according to task semantics, enabling multi-robot systems to address diverse semantic tasks that are difficult to formalize in advance \cite{roco,coela,icra}. However, text-only LLMs cannot directly perceive visual details and spatial relationships in physical environments, limiting their embodied understanding and decision-making capabilities. To address this limitation, VLMs jointly model visual and linguistic information \cite{llmva}, enabling robots to understand objects, regions, and their spatial relationships and to collaboratively execute open-ended, complex embodied tasks based on physical observations \cite{mcoconav,viki,AGrobot}. Nevertheless, existing methods still largely rely on centralized decision making, synchronized collaboration, and known map, and are often limited to either navigation or manipulation tasks, making it difficult to achieve efficient heterogeneous collaboration and closed-loop execution in unknown environments.

To enable VLM-driven heterogeneous multi-robot systems to execute complex tasks efficiently and reliably in unknown environments, the following key challenges must be addressed.
\textbf{1) Task Decomposition \& Assignment:} First, the system must accurately decompose complex instructions and assign subtasks according to the heterogeneous capabilities of different robots while maintaining robust planning.
\textbf{2) Efficient Information Sharing:} Second, heterogeneous collaboration in unknown environments requires efficient sharing of environmental observations and task-related information, whereas long-context prompts containing extensive observations and interaction histories may reduce reasoning accuracy and induce hallucinations.
\textbf{3) Heterogeneous Robot Collaboration:} Third, heterogeneous robots must effectively exploit their complementary strengths while resolving potential conflicts among concurrent subtasks as shown in Fig. \ref{fig1}.
\textbf{4) Unified Action Execution:} Finally, under a unified high-level decision framework, the system must generate robot-specific executable actions according to the capabilities of different robots without requiring separate training or fine-tuning for each robot type, thereby enabling heterogeneous robots to collaboratively accomplish complex tasks through closed-loop execution.

\begin{figure*}[t]
    \centering
    \includegraphics[width=6.4in]{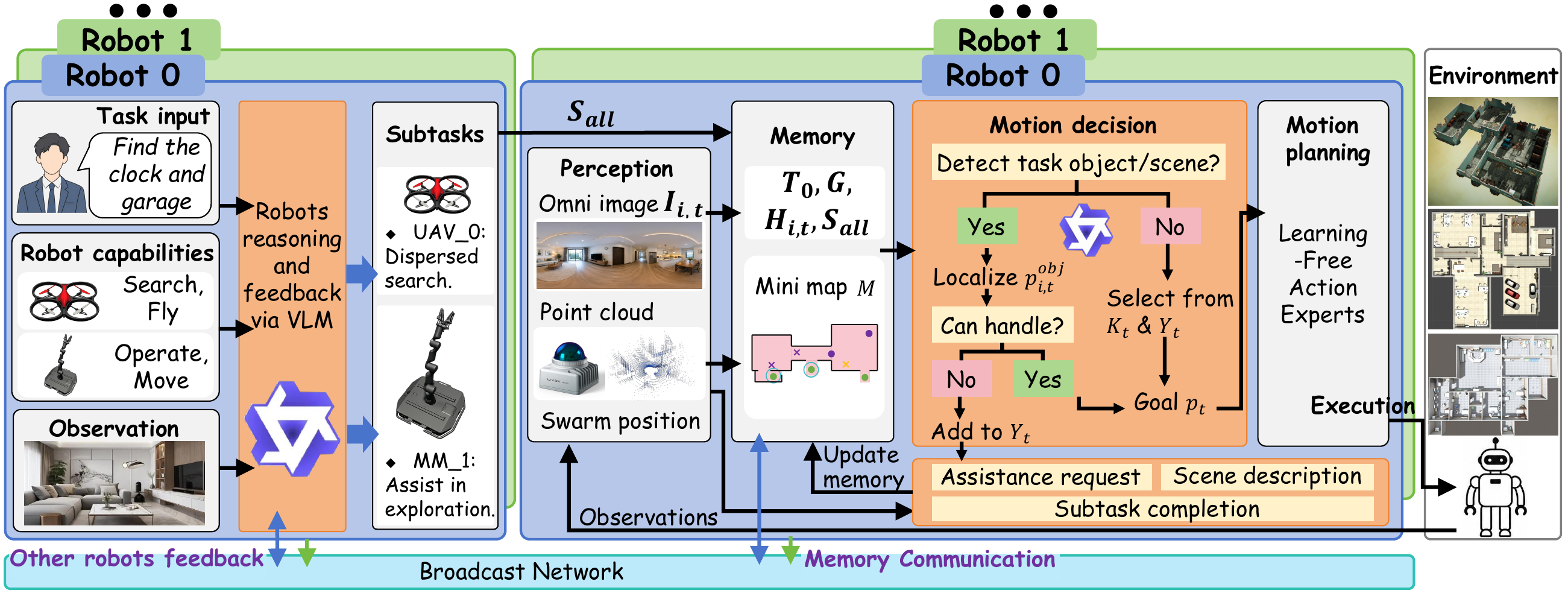}
    \caption{Overview of the proposed framework. Subtask is completed through shared decomposition and feedback, whereas perception, memory, exploration, assistance requests, and motion execution are performed asynchronously by each robot.}
    \label{fig:framework}
\end{figure*}

To address these challenges, as illustrated in Fig. \ref{fig:framework}, we propose a framework for heterogeneous embodied multi-robot collaboration in unknown environments. Each robot independently follows a decentralized and asynchronous perception--reasoning--action cycle while exchanging compact information on demand, avoiding centralized planning and synchronized decision rounds and reducing decision bottlenecks, delays, and context redundancy. First, we design a hybrid reasoning mechanism that combines robot capabilities and execution feedback to decompose, assign, and adapt complex instructions, thereby improving planning robustness. Second, each robot independently maintains a lightweight mini-map \cite{mini_map} and asynchronously exchanges compact language-based scene descriptions, local obstacle information, and exploration states. This design avoids reliance on predefined global maps or costly semantic maps while reducing communication, memory, and long-context reasoning burdens. Third, robots collaborate according to their complementary capabilities and dynamically request assistance from suitable teammates when necessary, thereby reducing conflicts among concurrent subtasks. Finally, the unified action decision pipeline enables the VLM to generate robot-specific high-level action decisions according to each robot's capabilities and feedback, which are then executed by corresponding learning-free action experts to perform exploration, long-range navigation, local visual understanding, and physical manipulation. We evaluate the framework in realistic simulated environments as shown in Fig. \ref{fig1}, including multi-room homes, indoor hospitals, and post-disaster ruins, using diverse ambiguous instructions and a heterogeneous team consisting of two aerial robots and one mobile manipulator. The results show that the proposed method generalizes across multiple VLMs without task- or robot-specific training or fine-tuning. All evaluated models achieve success rates above 70\% across different tasks, while the best-performing configuration reduces completion time by 55.8\% relative to the geometric greedy baseline.
In summary, the main contributions of this paper are as follows:
\begin{itemize}
\item We formulate a challenging decentralized embodied collaboration problem where heterogeneous agents interpret open-ended instructions, reason about complementary capabilities, and coordinate asynchronously in unknown environments.
\item We present D-VLC, a decentralized cognitive architecture that turns general-purpose VLMs into cooperative embodied agents through independent perception--reasoning--action loops.
\item We introduce capability-conditioned reasoning and bounded multimodal spatial memory, enabling grounded decisions, adaptive assistance, and embodiment-specific behaviors without task-specific training.
\item We evaluate D-VLC across diverse tasks, environments, embodiments, and VLM backbones, demonstrating generalization, visual grounding, capability awareness, and improved collaboration efficiency.
\end{itemize}

% \begin{itemize}
%     \item We propose a VLM-based framework for heterogeneous embodied multi-robot collaboration in unknown environments. Each robot reasons and acts independently while coordinating asynchronously, without centralized control or task-specific training.

%     \item We introduce a lightweight mini-map representation with hybrid exploration. It incrementally integrates swarm-wide spatio-temporal information to support interpretable collaborative exploration and navigation.

%     \item We develop a unified action decision mechanism that generates capability-aware, robot-specific high-level actions through a generic interface, which are executed by learning-free experts within a unified framework.

%     \item We conduct extensive experiments in different scenarios with diverse tasks and multiple VLM backbones, demonstrating the efficiency and generality of our framework.
% \end{itemize}

\section{Related Work}

\subsection{Vision-Language Reasoning and VLA Models}

Recent advances in VLMs have extended embodied agents from visual perception to semantic reasoning, spatial understanding, affordance prediction, and action-oriented decision-making. To improve grounded reasoning and reduce hallucinations, prior studies have explored chain-of-thought prompting, visual reasoning traces, tool-augmented perception, and reinforcement-learning-based reasoning optimization \cite{cot1,viscot,spatialr1,vtriune}. Meanwhile, VLA models such as RT-2 and OpenVLA directly map visual observations and language instructions to robot actions, showing promising generalization across manipulation tasks and embodiments \cite{rt2,openvla,vla_survey}. In parallel, VLN methods and navigation foundation models improve long-horizon navigation by aligning egocentric observations with natural-language instructions \cite{vln_survey,navfom}. However, most VLA/VLN methods focus on single-agent settings or task-specific navigation/manipulation, and rarely address decentralized coordination, heterogeneous capability matching, and low-memory global situational alignment in unknown large-scale scenes.

\subsection{Embodied Multi-Agent Collaboration}
Embodied multi-agent collaboration requires multiple physical agents to coordinate perception, communication, navigation, and action. Early LLM-based systems support task decomposition, inter-agent communication, and long-horizon planning \cite{smartllm,roco,coela}, but their lack of visual grounding limits reasoning about spatial constraints, object affordances, and fine-grained scene semantics. Recent VLM-based methods extend multi-robot systems to visual semantic navigation, decentralized exploration, and open-vocabulary object search \cite{conavgpt,mcoconav,dm3nav,goalvlm}. However, these methods primarily focus on navigation or object-goal search, with limited support for manipulation and heterogeneous capability matching. Heterogeneous systems such as VIKI-R \cite{viki} and COHERENT \cite{icra} further incorporate physical operation, but remain dependent on known environments or lack direct visual grounding for action-relevant decisions.
In summary, existing systems rarely support language understanding, visual grounding, navigation, physical operation, heterogeneous embodiments, and unknown-environment execution simultaneously. These limitations motivate a unified framework for efficient information alignment, heterogeneous capability collaboration, and closed-loop task execution.

\section{Method}

The system follows a hybrid architecture: task initialization uses shared decomposition and robot feedback, while all subsequent perception, decision, communication, navigation, and manipulation loops are executed locally. Thus, ``decentralized'' refers to online execution rather than to the one-time initialization stage. Figure~\ref{fig:framework} summarizes the information flow.

\subsection{System Setting and Information Flow}
Consider heterogeneous robots $\mathcal{N}=\{1,\ldots,n\}$ in an initially unknown environment. Given an ambiguous instruction $T_o$, the initialization module produces subtasks $G=\{g_1,\ldots,g_k\}$ from the swarm capability descriptions $S_{\mathrm{all}}=\{S_i\}_{i=1}^{n}$, where $S_i$ specifies sensing, mobility, and manipulation abilities. At time $t$, robot $i$ observes
\begin{equation}
O_{i,t}=\{I_{i,t},Z_{i,t},x_{i,t}\},
\end{equation}
where $I_{i,t}$ is a panoramic RGB observation, $Z_{i,t}$ contains geometric observations such as LiDAR points or a local occupancy grid, and $x_{i,t}$ is the robot pose. Its VLM context is
\begin{equation}
J_{i,t}=\{T_o,G,H_{i,t},S_{\mathrm{all}},M_{i,t}\},
\end{equation}
where $H_{i,t}$ stores compact task and communication history, and $M_{i,t}$ is the local copy of the shared abstract mini-map.

Robots do not transmit raw panoramic streams or complete point clouds to the VLMs of their teammates. Instead, they exchange structured updates
\begin{equation}
C_{i,t}=\{H_{o,t},M_{o,t},P_{o,t},K_{o,t},Y_{o,t},p^n_{o,t},f^{co}_{o,t}\}_{o\neq i},
\end{equation}
including task status, map elements, teammate positions, geometric candidates $K_{o,t}$, semantic candidates $Y_{o,t}$, selected exploration goals $p^n_{o,t}$, and assistance requests $f^{co}_{o,t}$. Each update is associated with a sender and time index. A robot merges the latest received state with its local observations; nearby candidate points are deduplicated, and a target already selected by a teammate is excluded unless no sufficiently separated alternative remains. Consequently, robot states need not be perfectly identical at every instant, but converge to a common abstract spatial state as updates arrive.

The high-level action is
\begin{equation}
A^H_{i,t}=\{m_{i,t},f^{td}_{i,t},f^{tc}_{i,t},f^{obj}_{i,t},
p^{obj}_{i,t},p^*_{i,t},f^{co}_{i,t}\},
\end{equation}
where $m_{i,t}\in\{\textit{explore},\textit{execute},\textit{request}\}$ denotes the current mode. The target passed to the learning-free motion layer is
\begin{equation}
p^*_{i,t}=
\begin{cases}
p^a_{i,t}, & m_{i,t}=\textit{execute},\\
p^n_{i,t}, & m_{i,t}=\textit{explore},\\
\emptyset, & m_{i,t}=\textit{request}.
\end{cases}
\end{equation}
The prompted VLM and the low-level expert are therefore separated as
\begin{equation}
A^H_{i,t}=\Pi_{\mathrm{VLM}}(J_{i,t},I_{i,t},K_t,Y_t), \qquad
\end{equation}
\begin{equation}
u_{i,t:t+\Delta t}=\Sigma_{\mathrm{LF}}^i(p^*_{i,t},Z_{i,t},x_{i,t},M_{i,t}).
\end{equation}
\subsection{Task Decomposition}
The initial ambiguous instruction is decomposed using the HMAS-2-style hybrid paradigm \cite{four,coela}. A robot first proposes subtasks from $T_o$, the initial observations $\{I_{i,0}\}$, and $S_{\mathrm{all}}$ via VLM. The proposal is returned to the other robots for reflection against their own embodiment descriptions and observations. A robot can reject an assignment that requires an unavailable sensor, mobility mode, or manipulation ability. The accepted subtask set $G$ is then broadcast once and stored together with the original instruction.

This initialization does not prescribe a complete long-horizon action sequence. During execution, each robot repeatedly grounds $T_o$ and $G$ against new visual evidence. Retaining $T_o$ prevents a mistaken or overly narrow subtask decomposition from becoming the only task representation, while the capability check constrains assignments that cannot be executed by the selected embodiment. After $G$ is established, no central module chooses exploration points, navigation goals, or manipulation actions.

\begin{figure}[t]
    \centering
    \includegraphics[width=3.3in]{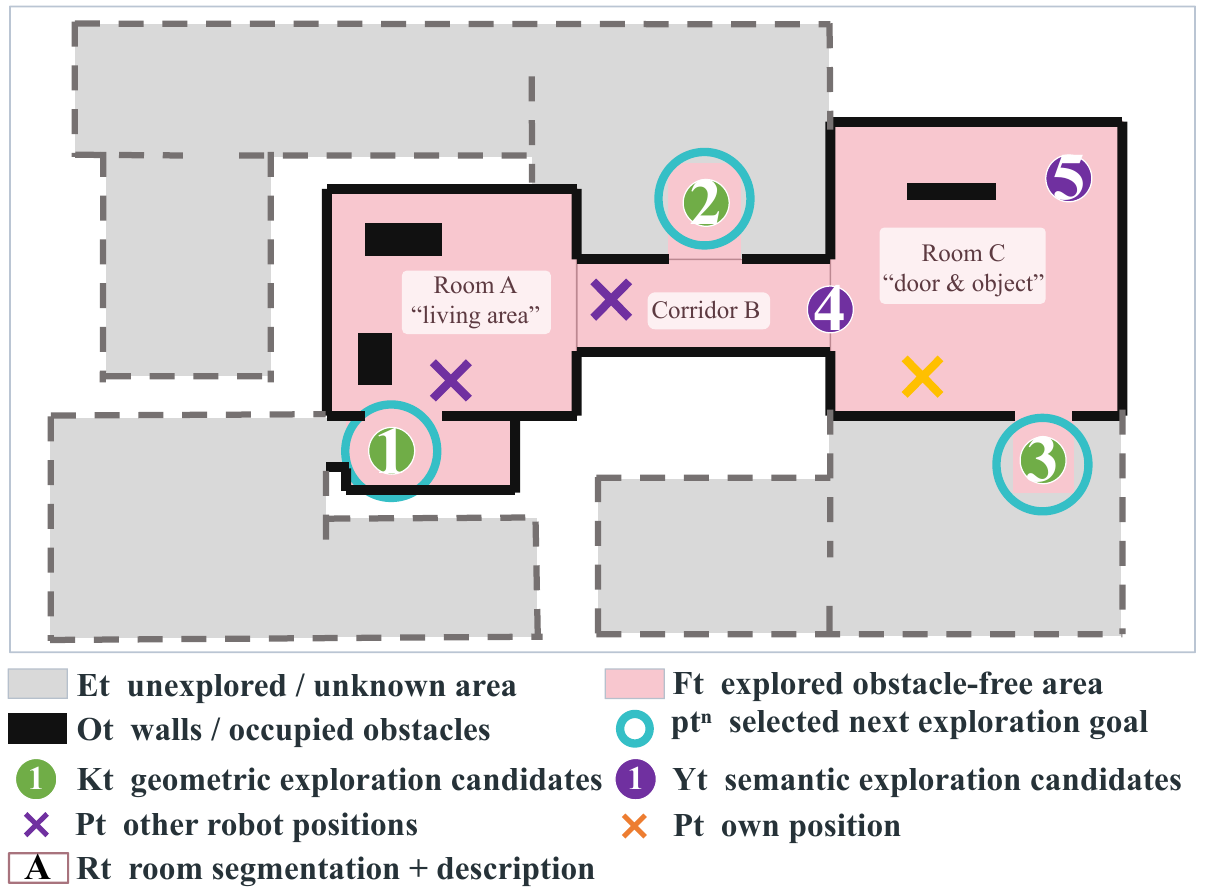}
    \caption{Mini-map as the VLM-readable spatial interface.}
    \label{fig:minimap}
\end{figure}

\subsection{Shared Spatial Memory}
For robot $i$, the memory state contains $T_o$, $G$, structured history $H_{i,t}$, swarm descriptions $S_{\mathrm{all}}$, and mini-map $M_{i,t}$. The mini-map is a $640\times480$ RGB representation inspired by navigation aids in open-world games \cite{mini_map}. Its spatial size is fixed throughout a trial, unlike a raw image history, growing point-cloud archive, or pixel-level mesh map \cite{icra,mcoconav}. Room semantics are stored as short language descriptions linked to segmented regions rather than as dense per-pixel labels.

The mini-map represents explored obstacle-free cells
$F_t=F_{t-1}\cup F^l_{i,t}\cup F^l_{o,t}$, occupied cells
$O_t=O_{t-1}\cup O^l_{i,t}\cup O^l_{o,t}$, room regions
$R_t=R_{i,t}\cup R_{o,t}$, unexplored space
$E_t=E_{t-1}\setminus(F_{i,t}\cup O_{i,t})$, robot positions
$P_t=P_{i,t}\cup P_{o,t}$, and selected exploration targets
$p_t^n=p^n_{i,t}\cup p^n_{o,t}$. Here, the superscript $l$ denotes locally observed geometry, while the subscript $o$ denotes information received from other robots. Free-space regions are rendered as compact color blocks, occupied cells form the obstacle boundary, and robot and target markers are drawn in separate visual channels so that they can be distinguished by the VLM.

Local free and occupied cells are projected from LiDAR-based mapping. During every local decision cycle, the panoramic observation is summarized into a concise description $d_{i,t}\in D_{i,t}$. Morphological segmentation \cite{room} divides the explored map into room-like regions, and the spatially nearest accumulated description is attached to each region. Consequently, the VLM can associate a coarse spatial region with phrases such as ``living area,'' ``corridor,'' or ``room containing a door,'' while the geometric layer remains independent of dense semantic reconstruction. The task history records only decision-relevant events, including discovered targets, completed subtasks, assistance requests, and teammate responses.

The memory also stores geometric candidates $K_t$ and semantic candidates $Y_t$. Teammate updates are fused into the same representation, allowing the VLM to reason about which regions have already been explored, where other robots are moving, and whether a candidate is already assigned. Communication is event-driven: map and scene updates are sent after a local exploration step, selected goals are broadcast when they change, and assistance messages are issued immediately after a capability mismatch. Receiving robots update their own $\mathcal{B}_{i,t+1}=\{T_o,G,H_{i,t+1},S_{\mathrm{all}},M_{i,t+1}\}$ and continue their local loop without waiting for a synchronized global planning round. We describe the representation as bounded rather than claiming measured communication or memory savings: the current experiments evaluate task behavior, while direct bandwidth and storage measurements remain future work.

\subsection{Exploration Cues}
The geometric candidate set is derived from the relationship between explored free space, occupied boundaries, room regions, and unknown space. For each room-like region $R_t^r$, occupied cells near each boundary are projected onto the boundary direction. Consecutive projected coordinates $q_j$ and $q_{j+1}$ define a raw opening when
\begin{equation}
g_{\min}\leq q_{j+1}-q_j\leq g_{\max},
\end{equation}
and the midpoint becomes a candidate $k_j$. Candidates caused by incomplete walls or map noise are rejected unless they are adjacent to observed free space outside the current room segmentation.

For a retained candidate, rays are traced over feasible angular sectors. A direction is valid when it reaches $E_t$ before intersecting $O_t$. If $r_{\mathrm{hit}}$ is the first distance to unknown space along the middle direction $\theta_{\mathrm{mid}}$ of a feasible sector, the navigation target is extended to
\begin{equation}
\tilde{k}_j=k_j+\alpha r_{\mathrm{hit}}
[\cos\theta_{\mathrm{mid}},\sin\theta_{\mathrm{mid}}]^\top,\qquad 0<\alpha<1.
\end{equation}
The factor $\alpha$ keeps the target in the transition area rather than directly inside unobserved space. Nearby candidates are merged locally and across robots:
\begin{equation}
K_t=\operatorname{Dedup}(K_{i,t}\cup K_{o,t}).
\end{equation}
A received candidate or selected target is retained only when it is sufficiently separated from existing ones, reducing duplicated exploration. The supplementary material gives the complete boundary, clustering, ray-extension, and filtering procedure.

Semantic candidates $Y_{i,t}$ are generated by VLM visual understanding. They include actual doors, corridor entrances, occlusions, and potential task-relevant objects that may not be recoverable from occupancy geometry alone. The shared set $Y_t=Y_{i,t}\cup Y_{o,t}$ is projected onto the same mini-map as $K_t$. The VLM therefore selects $p^n_{i,t}\in K_t\cup Y_t$ using task semantics, history, robot state, and spatial context rather than only Euclidean distance.

\subsection{VLM Motion Decision}
At time t, the VLM first determines whether the current panorama contains a task-related object or local scene. If so, a second grounded query identifies its image region. The detected region is associated with the closest LiDAR depth measurements within the bounding box. For regions without sufficient LiDAR coverage, the missing depth is further completed using MoGe-2 \cite{moge}. The recovered depth and the viewing direction of the grounded region are then combined with the current robot pose to estimate the global object position $p^{obj}_{i,t}$. A capability check then determines whether robot i can execute the required action. Executable targets become $p^a_{i,t}$; otherwise, the robot broadcasts an assistance request containing the semantic target, global location, and required capability, and preserves the location in $Y_t$.

If no task-relevant target is detected, the VLM selects an exploration cue from $K_t\cup Y_t$. The prompt provides the task, completed and pending subtasks, room descriptions, teammate locations, teammate goals, and the visualized candidate set. The output is constrained to a structured action mode and a candidate identifier rather than an unconstrained trajectory. Before execution, the selected point is checked against the latest teammate goals. If it conflicts with another target, the robot chooses a sufficiently separated alternative from the remaining candidates; if no alternative is available, it waits for the next map update instead of repeatedly selecting the same point.

After every decision, the robot updates its scene description, geometric state, task history, and selected target, and broadcasts the compact update. Task completion is checked against both the original instruction and the accumulated evidence from all robots. This prevents one robot from terminating the team after completing only its own subtask in a multi-target instruction. The supplementary material provides complete pseudocode for this loop.

\subsection{Motion Planning and Perception}
High-level task and exploration targets are converted into continuous, dynamically feasible actions by existing learning-free modules. A* \cite{astar} searches a long-range route on the mini-map. EGO Planner \cite{ego} generates local UAV trajectories, while TopAY \cite{topay} controls navigation and full pose manipulation for the mobile manipulator. FAST-LIO2 \cite{fastlio2} provides localization and local occupancy information from LiDAR. Omnidirectional RGB observations are used for VLM scene summarization, target detection, and grounding.

\section{Experiments}

\begin{figure*}[t]
    \centering
    \includegraphics[width=6.9in]{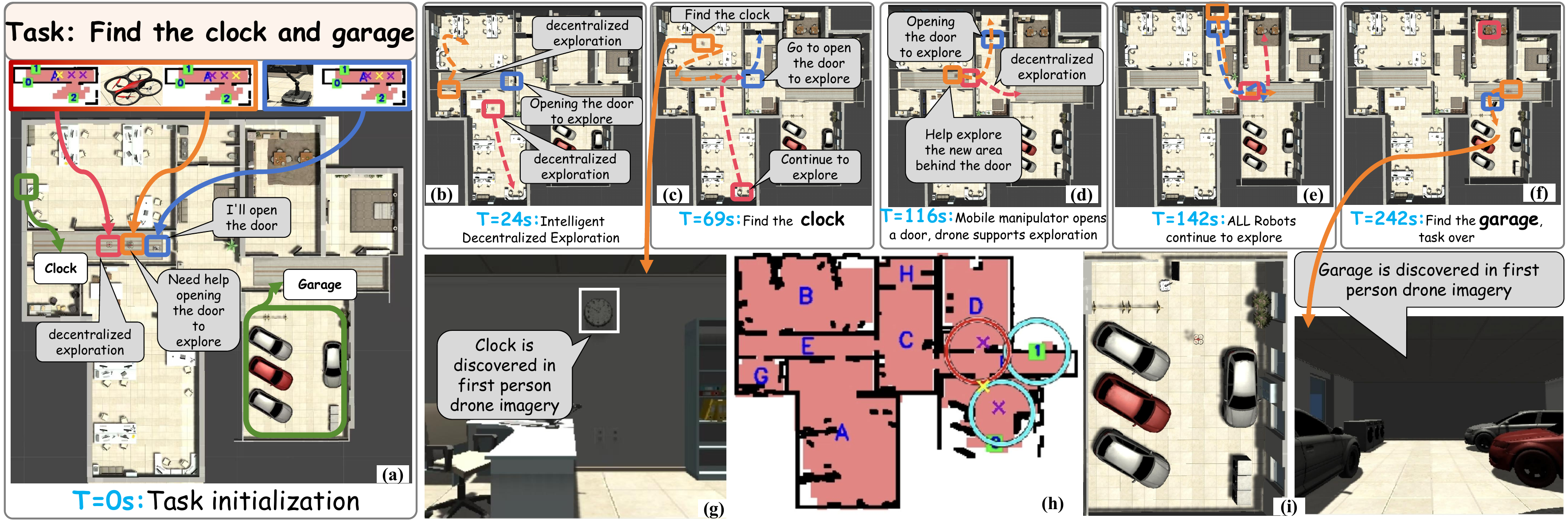}
    \caption{Representative home trial showing asynchronous exploration, clock detection, capability-aware door assistance, and garage discovery using first-person and mini-map evidence.}
    \label{task}
\end{figure*}

\subsection{Experimental Setup and Evaluation Protocol}
Experiments are conducted in Unity with ROS1. The main system is implemented in C++, with Python used only for VLM interaction, and runs on an AMD Ryzen~7 9700X CPU with approximately 32~GB RAM. The team comprises two UAVs for rapid aerial mobility and wide-area visual search and one mobile manipulator for ground navigation and physical interaction. This creates a clear capability asymmetry: UAVs efficiently inspect distant or elevated regions but cannot operate doors, whereas the mobile manipulator can interact with the environment but explores more slowly. More experimental details and related prompts can be found in the supplementary materials.

All methods share the same robot descriptions, perception modules, geometric and semantic candidate interface, motion planners, tasks, and VLM-based detection and grounding pipeline. Initial robot poses and instructions are fixed per scenario, while VLM outputs, exploration order, local planning, and interaction outcomes vary across trials. A trial succeeds only if every requested target is found and all required access or manipulation is completed within the run budget. We report success rate (SR), wall-clock completion time (CT), and high-level action steps for drone~0, drone~1, and the mobile manipulator ($SN_0$, $SN_1$, and $SN_2$); Total Steps is their sum. CT and step statistics use successful trials only, so SR measures reliability and the remaining metrics measure execution efficiency. The following experiments assess cross-scenario execution, backbone robustness under a controlled baseline, capability awareness, and visual grounding.

\begin{table*}[t]
\centering
\caption{Results over three task--scene pairs and 30 trials per method. CT and steps are mean $\pm$ standard deviation over successful trials; $\Delta$ is the reduction from Geometric Greedy.}
\label{tab:model_comparison}
\footnotesize
\setlength{\tabcolsep}{2.5pt}
\renewcommand{\arraystretch}{0.86}
\begin{tabular*}{\textwidth}{@{\extracolsep{\fill}}lcccccccc@{}}
\toprule
Method
& SR (\%)$\uparrow$
& CT (s)$\downarrow$
& $SN_0$
& $SN_1$
& $SN_2$
& Total Steps$\downarrow$
& $\Delta$CT (\%)$\uparrow$
& $\Delta$Steps (\%)$\uparrow$
\\
\midrule
Geometric Greedy
& 66.7
& 377.2$\pm$196.2
& 7.8$\pm$3.1
& 7.6$\pm$2.3
& 12.1$\pm$8.3
& 27.4$\pm$11.9
& --
& --
\\
Qwen3.5-Flash
& 86.7
& 200.5$\pm$60.5
& 9.3$\pm$3.8
& 8.8$\pm$1.9
& 3.9$\pm$1.5
& 22.0$\pm$5.3
& 46.9
& 19.9
\\
Doubao-Seed-2.1-Pro
& 80.0
& 198.6$\pm$35.8
& 5.3$\pm$1.9
& 5.7$\pm$1.2
& 4.0$\pm$1.2
& 15.0$\pm$2.6
& 47.4
& 45.1
\\
Claude-Opus-4-8
& \textbf{90.0}
& 184.9$\pm$22.3
& 6.9$\pm$1.0
& 6.4$\pm$1.6
& 3.6$\pm$1.2
& 16.9$\pm$3.0
& 51.0
& 38.5
\\
Gemini-3.5-Flash
& 83.3
& 368.5$\pm$88.7
& 5.6$\pm$1.7
& 4.8$\pm$1.7
& 5.0$\pm$1.6
& 15.4$\pm$3.9
& 2.3
& 43.7
\\
GPT-5.4-Mini
& 76.7
& \textbf{166.8$\pm$25.7}
& 5.5$\pm$1.2
& 5.5$\pm$0.8
& 2.4$\pm$0.8
& \textbf{13.3$\pm$2.2}
& \textbf{55.8}
& \textbf{51.3}
\\
\bottomrule
\end{tabular*}
\end{table*}

\subsection{Cross-Scenario Task Execution}
To evaluate the adaptability and generalization capability of
the proposed method across different scenarios and semantic
tasks, we test whether one framework can execute semantically different tasks in structurally different unknown environments without scene-specific maps, target locations, or scripted task policies. The three task--scene pairs are survivor search in post-disaster ruins, clock-and-garage search in a home, and toilet search in a hospital. They respectively emphasize dispersed exploration in irregular space, multi-target progress tracking with physical access through a closed door, and semantic search across multiple rooms. Together, they cover instruction understanding, visual grounding, long-range navigation, distributed exploration, and physical interaction while keeping the decision and execution interfaces unchanged.

Figure~\ref{task} shows the closed loop in a representative home trial. After task decomposition, the UAVs asynchronously search spatially separated regions while the mobile manipulator also explores. A UAV detects the clock and records it in the shared task state; because the garage remains pending, exploration continues. When a closed door blocks an unseen region, the UAV grounds the door, recognizes its manipulation mismatch, and requests assistance with the target location. The mobile manipulator opens the door, after which the team resumes distributed exploration and finds the garage.

This sequence is not predefined; it emerges from the instruction, accumulated visual evidence, shared mini-map, task history, and capability descriptions. The trial connects the main components: asynchronous exploration reduces waiting time, bounded spatial memory preserves team progress, visual grounding produces actionable targets, and capability-aware assistance transfers an infeasible action to a suitable embodiment.

\subsection{Backbone Robustness and Controlled Comparison}
For a controlled quantitative comparison, Geometric Greedy is implemented in the same robot stack and evaluated with five VLM backbones. When no task-related object is detected, it chooses the geometric candidate with the highest expected coverage. Detection and grounding, candidate generation, communication, capability checks, motion planning, low-level experts, and success criteria remain identical to D-VLC. Thus, the difference is geometry-only target selection versus selection conditioned on the task, scene descriptions, history, swarm states, and robot capabilities.

\begin{table}[t]
\centering
\caption{Capability coverage of representative systems.}
\label{tab0}
\renewcommand{\arraystretch}{0.82}
\setlength{\tabcolsep}{2.6pt}
\footnotesize
\begin{tabular}{lcccccc}
\hline
Methods & L. & V. & Nav. & Ope. & H.E. & U.E. \\
\hline
RoCo \smallcite{roco}      & \tikz[baseline=-0.6ex]{\draw (0,0) circle (0.8ex);} &  $\times$       &     $\times$    & \tikz[baseline=-0.6ex]{\draw (0,0) circle (0.8ex);} &   $\times$    &   $\times$   \\
Co-ELA \smallcite{coela}   & \tikz[baseline=-0.6ex]{\draw (0,0) circle (0.8ex);} &  $\times$    & \tikz[baseline=-0.6ex]{\draw (0,0) circle (0.8ex);} & \tikz[baseline=-0.6ex]{\draw (0,0) circle (0.8ex);} &  $\times$     &      $\times$  \\
MCoCoNav \smallcite{mcoconav} & \tikz[baseline=-0.6ex]{\draw (0,0) circle (0.8ex);} & \tikz[baseline=-0.6ex]{\draw (0,0) circle (0.8ex);} & \tikz[baseline=-0.6ex]{\draw (0,0) circle (0.8ex);} &  $\times$   &    $\times$   & \tikz[baseline=-0.6ex]{\draw (0,0) circle (0.8ex);} \\
Co-NavGPT\\[-1mm]\smallcite{conavgpt} & \tikz[baseline=-0.6ex]{\draw (0,0) circle (0.8ex);} & \tikz[baseline=-0.6ex]{\draw (0,0) circle (0.8ex);} & \tikz[baseline=-0.6ex]{\draw (0,0) circle (0.8ex);} &  $\times$   &  $\times$     & \tikz[baseline=-0.6ex]{\draw (0,0) circle (0.8ex);} \\
VIKI-R \smallcite{viki}   & \tikz[baseline=-0.6ex]{\draw (0,0) circle (0.8ex);} & \tikz[baseline=-0.6ex]{\draw (0,0) circle (0.8ex);} & \tikz[baseline=-0.6ex]{\draw (0,0) circle (0.8ex);} & \tikz[baseline=-0.6ex]{\draw (0,0) circle (0.8ex);} & \tikz[baseline=-0.6ex]{\draw (0,0) circle (0.8ex);} &  $\times$   \\
COHERENT \smallcite{icra} & \tikz[baseline=-0.6ex]{\draw (0,0) circle (0.8ex);} &  $\times$   & \tikz[baseline=-0.6ex]{\draw (0,0) circle (0.8ex);} & \tikz[baseline=-0.6ex]{\draw (0,0) circle (0.8ex);} & \tikz[baseline=-0.6ex]{\draw (0,0) circle (0.8ex);} & \tikz[baseline=-0.6ex]{\draw (0,0) circle (0.8ex);} \\
\hline
\textbf{Ours}      & \tikz[baseline=-0.6ex]{\draw[line width=0.8pt] (0,0) circle (0.8ex);} & \tikz[baseline=-0.6ex]{\draw[line width=0.8pt] (0,0) circle (0.8ex);} & \tikz[baseline=-0.6ex]{\draw[line width=0.8pt] (0,0) circle (0.8ex);} & \tikz[baseline=-0.6ex]{\draw[line width=0.8pt] (0,0) circle (0.8ex);} & \tikz[baseline=-0.6ex]{\draw[line width=0.8pt] (0,0) circle (0.8ex);} & \tikz[baseline=-0.6ex]{\draw[line width=0.8pt] (0,0) circle (0.8ex);} \\
\hline
\end{tabular}
\end{table}

Table~\ref{tab:model_comparison} shows that all VLM-guided variants raise SR from 66.7\% to 76.7--90.0\%. Claude-Opus-4-8 has the highest SR (90.0\%), while GPT-5.4-Mini has the shortest CT and fewest steps, reducing them by 55.8\% and 51.3\%. The backbones therefore exhibit a reliability--efficiency trade-off: Gemini-3.5-Flash, for example, reduces steps by 43.7\% but CT by only 2.3\%, showing that fewer high-level decisions need not yield shorter wall-clock execution.
Per-robot counts explain part of the gain. Relative to Greedy, $SN_2$ falls from 12.1 to 2.4--5.0 for every backbone, whereas UAV counts vary less consistently. Task- and capability-conditioned reasoning therefore avoids assigning routine exploration repeatedly to the slower mobile manipulator and reserves it for interaction-dependent subtasks. Overall, the interface transfers across backbones, and semantic team-conditioned exploration is more reliable and generally more efficient than geometry-only selection under the same perception and control stack.

\subsection{Component Analysis and Ablation Study}
Existing embodied multi-robot systems typically focus on specific capabilities, such as language reasoning, navigation, or manipulation. However, our setting requires integrated language understanding, visual grounding, exploration, heterogeneous coordination, and physical execution, making direct comparison with existing methods difficult.

To analyze the capability gap, we summarize representative embodied multi-robot systems in Table~\ref{tab0}: language understanding (L.), visual perception (V.), navigation (Nav.), physical operation (Ope.), heterogeneous embodiments (H.E.), and unknown environments (U.E.). Existing approaches generally support only subsets of these capabilities: LLM-based methods lack visual grounding, VLM-based methods mainly target semantic navigation, and physical collaboration methods are often restricted to predefined scenarios. In contrast, our framework unifies these capabilities in a zero-shot manner. Furthermore, comparisons with methods using known global maps are not directly fair, as they exploit environmental structures and target priors unavailable in our unknown-environment setting.

To further validate the necessity of these capabilities, we design targeted physical interaction tasks with heterogeneous robots, including appearance-based grasping, relation-based grasping, and clock reading. These tasks require visual grounding, semantic reasoning, navigation, and manipulation simultaneously. The robot team must identify targets from language instructions, distinguish visually similar objects, and request appropriate robots when manipulation is required. Results show that text-only LLM-based methods fail in tasks requiring fine-grained visual understanding, such as clock reading, due to the lack of visual grounding. In contrast, our VLM-driven framework successfully grounds targets and generates executable actions through heterogeneous collaboration.
These results demonstrate that existing methods lack essential components for open-world heterogeneous embodied collaboration, while our framework provides a unified solution integrating language understanding, visual grounding, exploration, coordination, and physical execution.

\begin{figure}[t]
    \centering
    \includegraphics[width=3.3in]{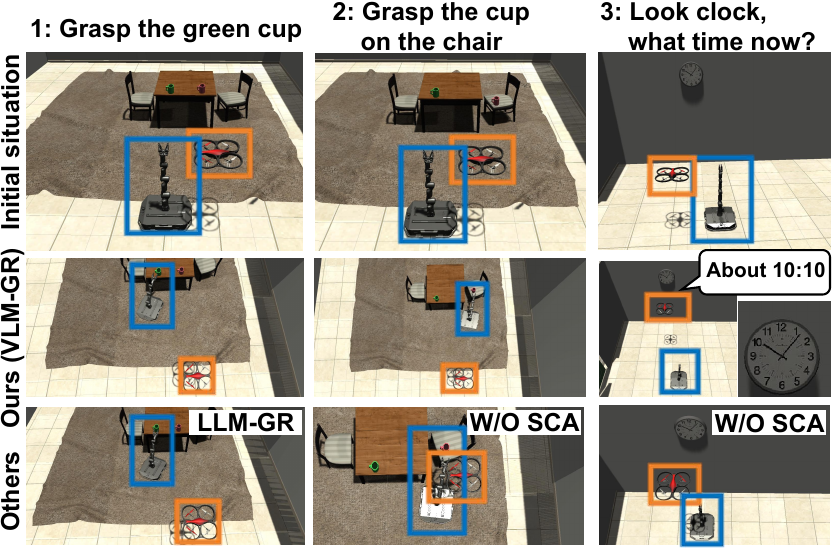}
    \caption{Component analysis of capability awareness and visual grounding.}
    \label{task00}
\end{figure}

\begin{table}[t]
\centering
\caption{Diagnostics for appearance grasping (Task 1), relational grasping (Task 2), and clock reading (Task 3).}
\label{tab1}
\setlength{\tabcolsep}{12pt}
\renewcommand{\arraystretch}{0.60}
\footnotesize
\begin{tabular}{lccc}
\toprule
Method & Task 1 & Task 2 & Task 3 \\
\midrule
Ours w/o SCA & $\times$ & $\times$ & $\checkmark$ \\
Ours (LLM-GR) & $\checkmark$ & $\checkmark$ & $\times$ \\
\midrule
\textbf{Ours (VLM-GR)} & $\checkmark$ & $\checkmark$ & $\checkmark$ \\
\bottomrule
\end{tabular}
\end{table}

% Table~\ref{tab1} and Fig.~\ref{task00} show complementary failures. Without SCA, both manipulation tasks fail because the reasoner cannot match the required operation to a suitable embodiment. LLM-GR completes them when textual target descriptions suffice, but fails clock reading: it can approach the clock yet cannot determine its displayed time without direct visual evidence. Full VLM-GR completes all three tasks.

% The components play distinct roles. Capability conditioning determines \emph{which robot should act}, preventing physically infeasible assignments, while visual grounding determines \emph{what is present and what state it is in} for appearance-, relation-, and state-dependent decisions. Together they connect semantic reasoning to embodiment-specific execution rather than treating allocation and visual understanding independently.

\section{Conclusion}
We presented a heterogeneous embodied multi-robot framework for ambiguous instruction execution in unknown environments. It combines shared task initialization with decentralized asynchronous VLM reasoning, a bounded mini-map, geometric--semantic exploration, capability-aware assistance, and a unified action-decision pipeline. Experiments in three simulated scenes with multiple VLM backbones demonstrate cross-scenario generalization and substantially lower successful-run completion time and action steps than a greedy baseline. Future work will extend the framework to physical robots, larger teams, communication-constrained settings, and lightweight on-device VLMs.

\bibliography{references}

\appendix
\section{Appendix}
\section{Geometric Exploration Candidate Extraction}
\label{app:geometric_candidate_extraction}
This appendix details the computation of the geometric exploration candidates \(K_{i,t}\) used in the mini-map memory. In the implementation, these candidates are stored as \texttt{door} objects. However, they should not be interpreted as semantic doors or physically operable doors. Instead, they correspond to geometry-induced frontier candidates on the boundary of the currently explored space. Therefore, they belong to the geometric exploration candidate set \(K_{i,t}\), rather than the semantic exploration candidate set \(Y_{i,t}\). The latter is produced by VLM-based visual understanding from panoramic observations, such as semantic doors, corridor entrances, occlusions, or task-relevant objects.

At time \(t\), robot \(i\) maintains a mini-map \(M_{i,t}\), which encodes the explored obstacle-free area \(F_t\), occupied obstacles \(O_t\), unexplored area \(E_t\), room segmentation and descriptions \(R_t\), robot positions \(P_t\), geometric exploration candidates \(K_t\), semantic exploration candidates \(Y_t\), and selected next exploration goals \(p_t^n\). The following procedure computes the local geometric candidate set \(K_{i,t}\) from the geometric elements \((F_t,O_t,E_t,R_t)\) represented in \(M_{i,t}\).

For each room-like region \(R_t^r\in R_t\), where \(r\) indexes the room region, we approximate its boundary by a rectangle
\[
R_t^r=[x_{\min}^r,x_{\max}^r]\times[y_{\min}^r,y_{\max}^r].
\]
The four boundaries of \(R_t^r\) are examined independently. Taking the upper horizontal boundary as an example, occupied cells near this boundary are projected onto the boundary direction:
\[
W_{\mathrm{top}}^r =
\left\{
x \mid (x,y)\in O_t,\ 
x_{\min}^r\leq x\leq x_{\max}^r,\ 
|y-y_{\min}^r|<\tau_w
\right\},
\]
where \(\tau_w\) is the wall-thickness tolerance. After sorting the projected occupied coordinates,
\[
x_1 < x_2 < \cdots < x_k,
\]
the algorithm searches for discontinuities between adjacent occupied cells:
\[
g_j=x_{j+1}-x_j.
\]
If the gap satisfies
\[
g_{\min}\leq g_j\leq g_{\max},
\]
then the midpoint of this gap is regarded as a raw geometric exploration candidate:
\[
k_j=\left(\frac{x_j+x_{j+1}}{2},\ y_{\min}^r\right).
\]
The same procedure is applied to the lower horizontal boundary and to the two vertical boundaries. For vertical boundaries, occupied cells are projected along the \(y\)-direction, and the candidate position is computed from the midpoint of the detected vertical gap. This step converts two-dimensional boundary opening detection into one-dimensional discontinuity detection along room boundaries.

To suppress false candidates caused by map noise or incomplete local geometry, the algorithm further checks whether the candidate is adjacent to exterior observed free space. Specifically, observed free cells inside the current room segmentation are removed:
\[
F_t^{\mathrm{out}}
=
F_t\setminus \bigcup_{R_t^r\in R_t} R_t^r .
\]
The set \(F_t^{\mathrm{out}}\) contains observed free cells that lie outside the segmented room regions. These cells are clustered according to spatial proximity. Two cells \(p_a,p_b\in F_t^{\mathrm{out}}\) are considered connected if
\[
\|p_a-p_b\|_2 \leq \tau_c,
\]
which produces exterior free-space clusters
\[
\mathcal{C}_t=\{C_t^1,C_t^2,\ldots,C_t^m\}.
\]
For a raw candidate \(k_j\), if there exists an exterior free-space cluster sufficiently close to it,
\[
\min_{C_t^q\in \mathcal{C}_t}\min_{p\in C_t^q}
\|k_j-p\|_2
\leq \tau_p,
\]
then \(k_j\) is regarded as an exterior-free-space-adjacent geometric candidate. Such a candidate is more likely to correspond to a valid frontier entrance toward unexplored space, rather than a spurious gap caused by local map artifacts.

For exterior-free-space-adjacent candidates, a ray-based extension is further applied to generate safer and more informative navigation targets. Let \(k_j\) be the ray origin. The algorithm samples directions over \(360^\circ\):
\[
\theta_l=\frac{2\pi l}{N_r},\quad l=0,1,\ldots,N_r-1,
\]
and traces rays
\[
p_l(r)
=
k_j+r
\begin{bmatrix}
\cos\theta_l\\
\sin\theta_l
\end{bmatrix},
\quad 0\leq r\leq r_{\max}.
\]
If a ray reaches the unexplored region \(E_t\) before hitting occupied cells in \(O_t\), this direction is treated as a feasible frontier direction. The algorithm then groups consecutive feasible rays into angular sectors and selects the middle direction \(\theta_{\mathrm{mid}}\) of each valid sector as the representative direction. Let \(r_{\mathrm{hit}}\) denote the distance at which this representative ray first reaches \(E_t\). The extended candidate is computed as
\[
\tilde{k}_j
=
k_j+
\alpha r_{\mathrm{hit}}
\begin{bmatrix}
\cos\theta_{\mathrm{mid}}\\
\sin\theta_{\mathrm{mid}}
\end{bmatrix},
\quad 0<\alpha<1.
\]
Since \(\alpha<1\), the generated target is not placed directly inside the unexplored region, but in the transition area between observed free space \(F_t\) and unexplored space \(E_t\). This improves navigation safety while still guiding the robot toward informative unknown regions.

After collecting all raw and extended candidates, nearby candidates are merged to remove duplicates. Let \(\bar{K}_{i,t}\) denote the set of candidates generated by robot \(i\) before deduplication. Two candidates \(k_a,k_b\in\bar{K}_{i,t}\) are considered duplicates if
\[
\operatorname{dist}(k_a,k_b)<\tau_m,
\]
where \(\operatorname{dist}(\cdot,\cdot)\) can be implemented as either Euclidean distance or grid-based distance on the mini-map. Duplicates are merged or represented by a single candidate. In addition, candidates that only connect two already known room regions are filtered out, because they do not correspond to frontiers leading to unexplored areas. The remaining set forms the local geometric exploration candidates of robot \(i\):
\[
K_{i,t}=\operatorname{Dedup}(\bar{K}_{i,t}).
\]

In the distributed multi-robot setting, each robot receives geometric exploration candidates from other robots. Let \(K_{o,t}\) denote the set of candidates received from other robots. Robot \(i\) fuses its local candidates with the received candidates by distance-based duplicate removal:
\[
K_t
=
\operatorname{Dedup}\left(K_{i,t}\cup K_{o,t}\right).
\]
Equivalently, a received candidate \(k^o\in K_{o,t}\) is inserted into the local candidate set only if it is sufficiently far from all existing local candidates:
\[
K_{i,t}
\leftarrow
K_{i,t}
\cup
\left\{
k^o\in K_{o,t}
\mid
\min_{k\in K_{i,t}}\operatorname{dist}(k^o,k)>\tau_{\mathrm{dist}}
\right\}.
\]
The fused set \(K_t\) is drawn on the mini-map \(M_{i,t}\) and used together with the semantic exploration candidates \(Y_t\) for subsequent motion decision-making. When no task-relevant object or scene is detected, the VLM-based motion decision module selects the next exploration goal from these candidate sets:
\[
p_{i,t}^n \in K_t \cup Y_t.
\]
According to the high-level action formulation, when no task-relevant target is detected, the target sent to the low-level planner is
\[
p_{i,t}^{*}=p_{i,t}^{n}.
\]
Thus, the proposed geometric extraction procedure provides the candidate set \(K_{i,t}\), which is fused into \(K_t\) and eventually supports decentralized VLM-based exploration planning.

% Algorithm~\ref{alg:vlm_motion_decision} is included in the main paper.
\section{VLM Motion Decision}

Algorithm~\ref{alg:vlm_motion_decision} summarizes the loop of VLM motion decision.
 
\begin{algorithm}[t]
\caption{VLM-Based Local Decision Loop}
\label{alg:vlm_motion_decision}
\begin{algorithmic}[1]
\REQUIRE $J_{i,t},I_{i,t},Z_{i,t},x_{i,t},K_t,Y_t$
\ENSURE $A^H_{i,t},p^*_{i,t},\mathcal{B}_{i,t+1}$
\STATE $(f^{td},f^{obj})\leftarrow\textsc{DetectObj}(J_{i,t},I_{i,t})$
\IF{$f^{td}=1$}
  \STATE $p^{obj}\leftarrow\textsc{GroundPose}(I_{i,t},Z_{i,t},x_{i,t},f^{obj})$
  \IF{$\textsc{CanHandle}(i,f^{obj})$}
    \STATE $m\leftarrow\textit{execute}$; $p^*\leftarrow\textsc{MakeGoal}(p^{obj},f^{obj})$
  \ELSE
    \STATE $m\leftarrow\textit{request}$; $f^{co}\leftarrow\textsc{AskAssist}(f^{obj},p^{obj},S_{\mathrm{all}})$
    \STATE $Y_t\leftarrow Y_t\cup\{p^{obj}\}$
  \ENDIF
\ELSE
  \STATE $m\leftarrow\textit{explore}$; $p^*\leftarrow\textsc{PickPoint}(J_{i,t},K_t,Y_t)$
\ENDIF
\STATE $(H_{i,t+1},M_{i,t+1})\leftarrow\textsc{UpdateMem}(\cdot)$
\STATE $\textsc{Broadcast}(H_{i,t+1},M_{i,t+1},f^{co})$
\STATE $f^{tc}\leftarrow\textsc{CheckDone}(T_o,G,H_{i,t+1})$
\RETURN $A^H_{i,t},p^*_{i,t},\mathcal{B}_{i,t+1}$
\end{algorithmic}
\end{algorithm}

\section{Additional Experimental Results}
\begin{figure*}[!t]
    \centering

    \includegraphics[width=7.0in]{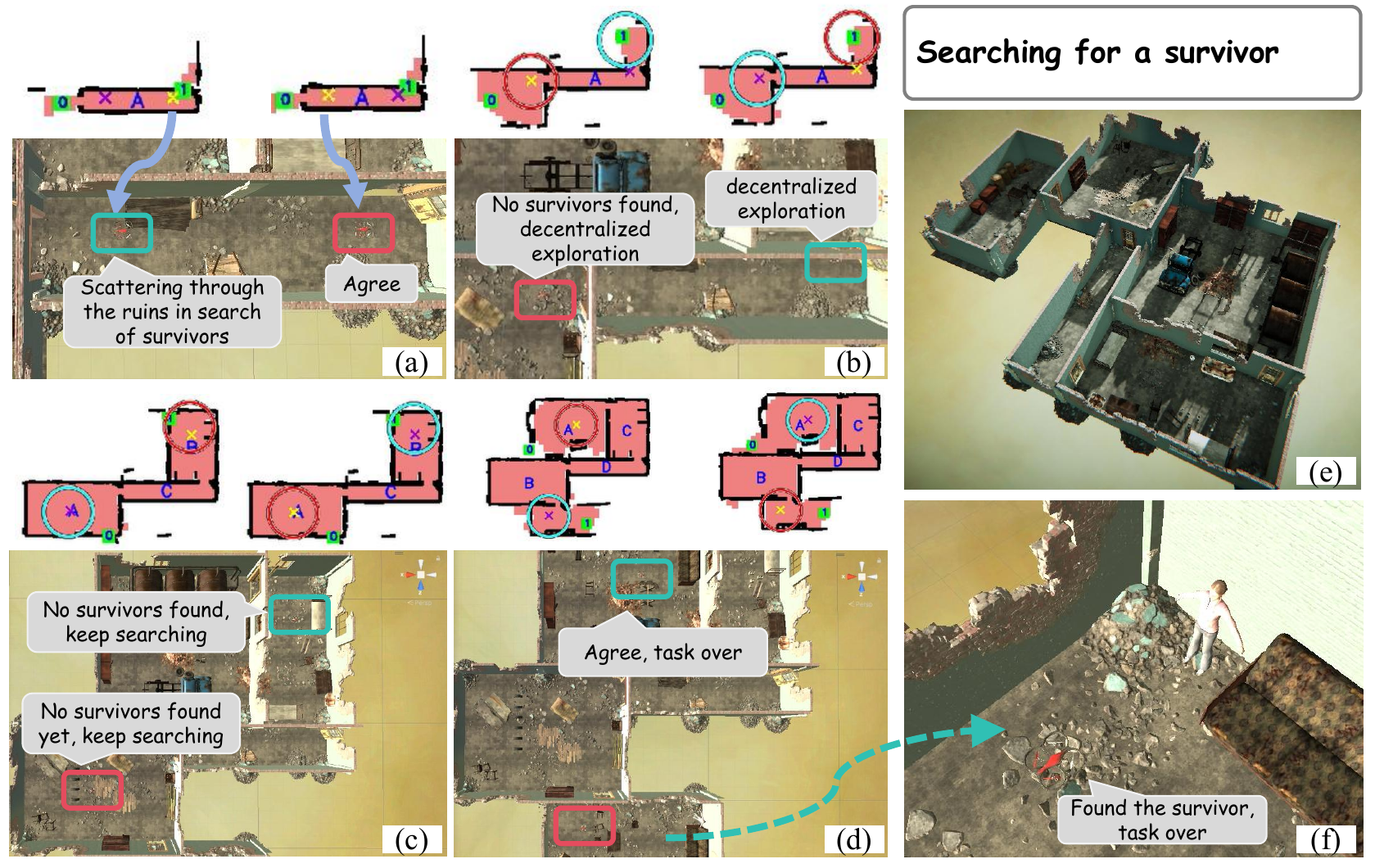}
    \caption{Qualitative example of multi-robot execution in the post-disaster ruins scenario. The snapshots illustrate distributed exploration, swarm communication, and task grounding for survivor search.}
    \label{fig:task_scenarios0}
\end{figure*}

\begin{figure*}[!t]
    \centering
    \includegraphics[width=7.0in]{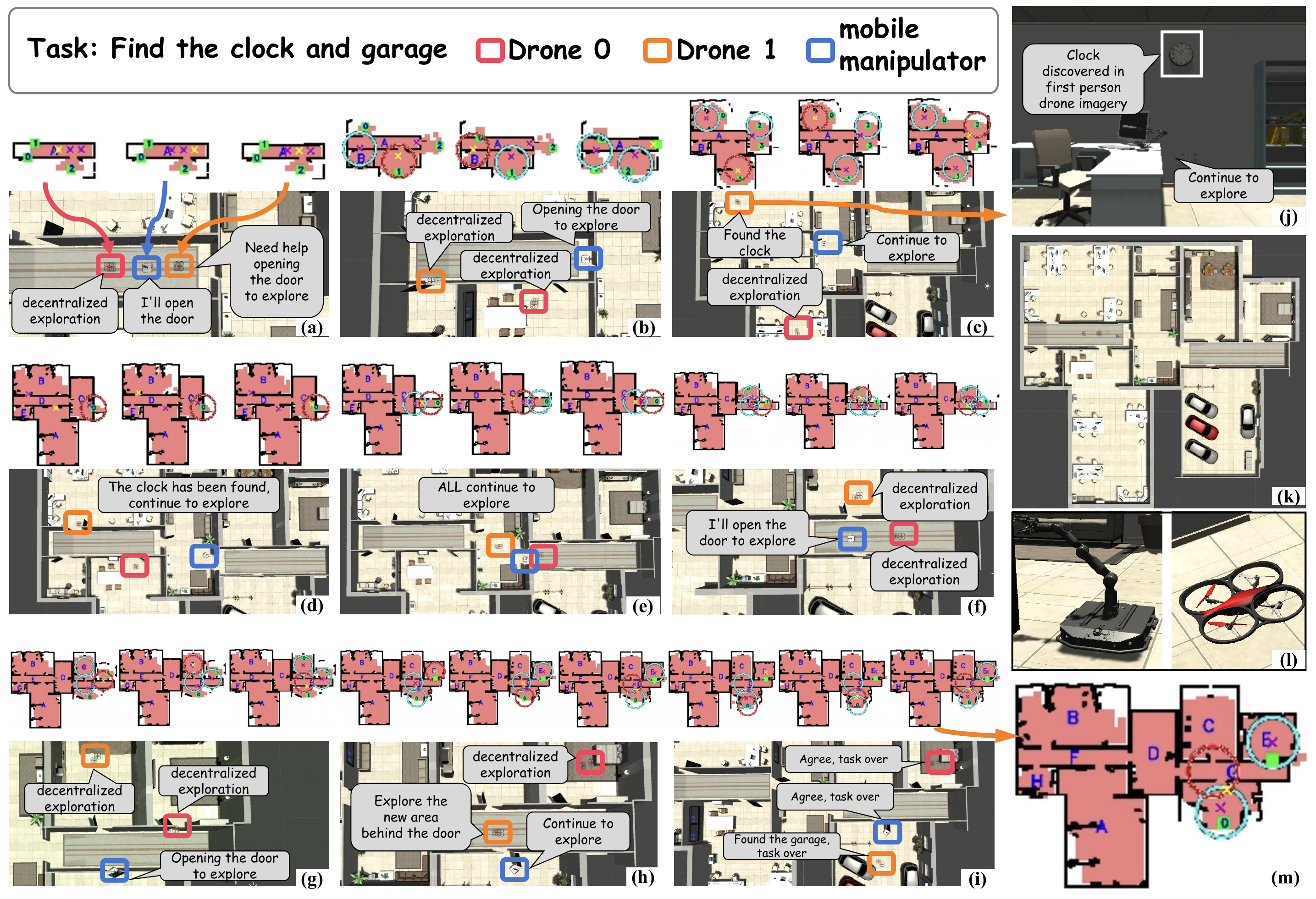}
    \caption{
    Qualitative example of multi-robot execution in the home.
    }
    \label{fig:home_additional}
\end{figure*}

\begin{figure*}[!t]
    \centering

    \includegraphics[width=7.0in]{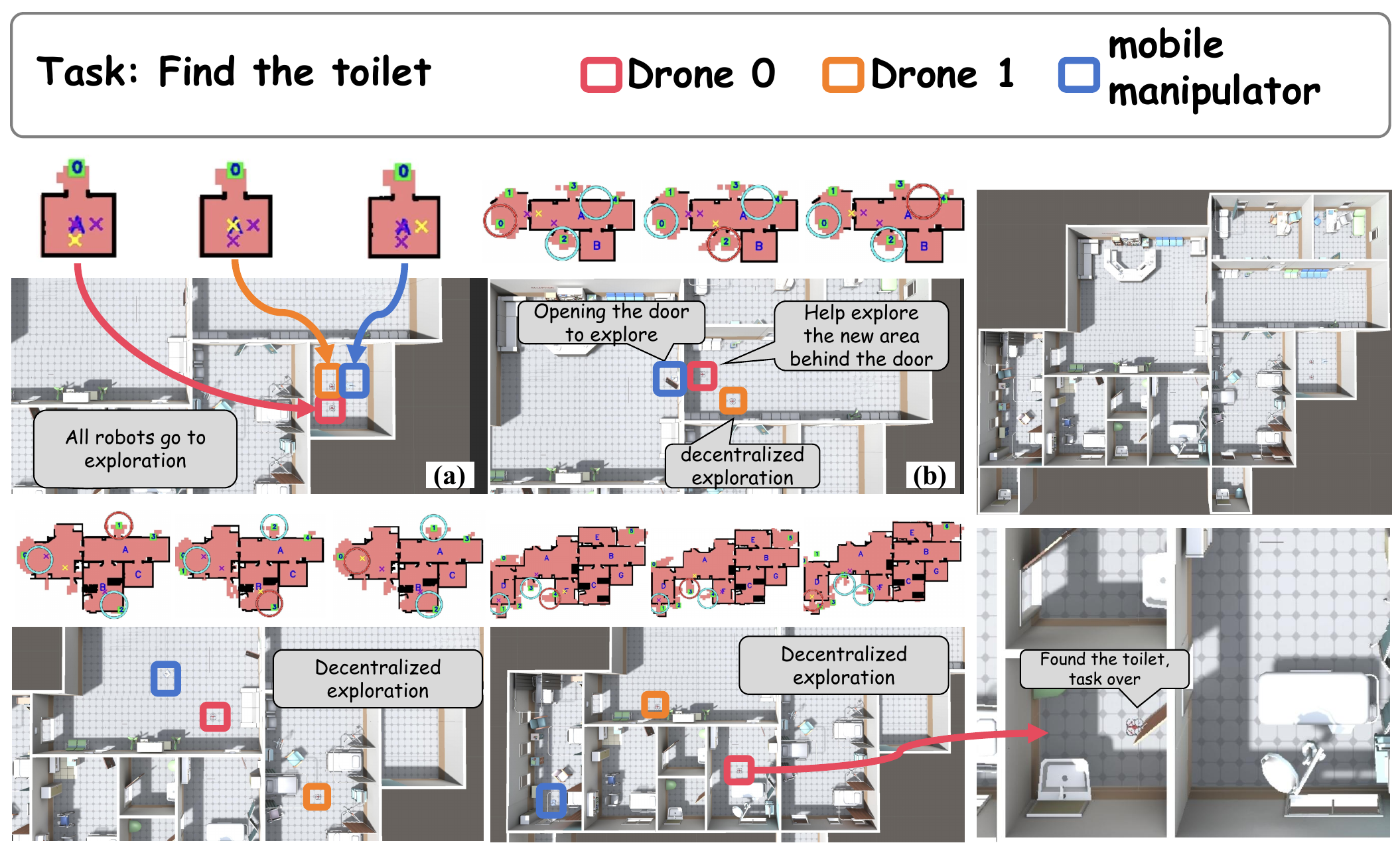}
    \caption{Qualitative example of multi-robot execution in the hospital scenario.}
    \label{fig:task_scenarios}
\end{figure*}

Additional qualitative results are shown in Figs.~\ref{fig:task_scenarios0}, \ref{fig:home_additional}, and~\ref{fig:task_scenarios}.

% \subsection{Experimental Protocol and Reporting}
% For every method, the same three task definitions, robot embodiments, initial scenario settings, motion planners, and structured decision interface are used. Each task is repeated six times, giving 18 trials per method. SR is computed over all trials, whereas CT and the step statistics are computed over successful trials only. The greedy baseline replaces VLM-based mini-map candidate reasoning with a distance-based point-selection rule while retaining the remaining perception and motion stack. It should therefore be interpreted as an integrated baseline rather than as a complete component-wise ablation.

\subsection{Representative Failure Cases}
The recorded runs exhibit four recurring failure stages: (1) a requested object, doorway, or relevant scene is missed or incorrectly classified; (2) a correct semantic target is grounded to an inaccurate image region or depth; (3) map changes or target conflicts temporarily leave no valid exploration or navigation target; and (4) navigation, door approach, or manipulation times out or enters a replanning loop. These qualitative cases are included to clarify the difference between successful-run efficiency and overall task reliability. A complete quantitative failure taxonomy is left for future evaluation because the current trial records were not uniformly annotated at that level.

\subsection{Failure Analysis and Limitations}
Observed failures occur at several stages. The VLM may miss or misclassify a requested object or doorway, producing repeated exploration or an incorrect completion judgment. A correct semantic detection may still be grounded to an inaccurate image region or depth, yielding an invalid global target. Map changes after door opening can temporarily invalidate an exploration candidate or global path. Finally, local navigation and manipulation can time out or enter a replanning loop even when the high-level target is correct. These cases explain why shorter trajectories do not always correspond to higher SR and identify perception grounding and execution recovery as the main remaining bottlenecks.

The study is limited to simulation, three robots, three task categories, and an integrated greedy baseline. It does not isolate every framework component or measure communication traffic under packet loss and latency. In addition, task initialization uses a shared consensus stage, although online execution is decentralized. We state these boundaries explicitly and leave physical-robot deployment, larger teams, network stress tests, and controlled component ablations to future work.

\section{Prompt Design and Implementation}
\label{app:prompt-design}

\subsection{Prompt Construction and Runtime Interface}\label{prompt-construction-and-runtime-interface}

Each model request consists of a fixed task-specific system prompt, a runtime-generated user prompt, and, where required, one or more visual observations.

The scene-graph and exploration modules assemble task state as JSON. This state may contain the general mission, robot identity, candidate exploration targets, teammate goals, action history, current observations, or task-allocation dialog. The JSON is transported through the ROS PromptMsg.prompt field. The Python interface uses PromptMsg.prompt\_type to select the corresponding system prompt and, for visual tasks, attaches the latest local map or camera observation. Model outputs are constrained to compact JSON objects so that the downstream C++ module can convert language-model inference into navigation, grounding, collaboration, or termination decisions.

The variants numbered A, AA, and B are camera bindings rather than independently designed prompts. Specifically, A/A1/A2/A3 and B/B1/B2/B3 share their respective templates while receiving different camera topics. AA/AA1/AA2/AA3 share the same multi-view verification instruction; the interface associates each variant with a photo index. If a required image is unavailable or image encoding fails, the interface falls back to a text-only request. This fallback is an implementation safeguard and does not change the system-prompt definition.

\subsection{Prompt Inventory and Functional Mapping}\label{prompt-inventory-and-functional-mapping}

\begin{table*}[htbp]
\centering
\caption{Prompt inventory and functional mapping. Camera-indexed variants sharing the same semantic template are grouped in one row.}
\label{tab:prompt-inventory}
\scriptsize
\setlength{\tabcolsep}{3pt}
\renewcommand{\arraystretch}{1.12}
\begin{tabularx}{\textwidth}{@{}P{0.205\textwidth}P{0.225\textwidth}P{0.245\textwidth}Y@{}}
\toprule
\textbf{Function} & \textbf{\texttt{PromptMsg} type} & \textbf{Input modality} & \textbf{Principal response fields} \\
\midrule
Room semantic classification & \path{PROMPT_TYPE_ROOM_PREDICTION} & JSON text & \path{results[].id}, \path{description}, \path{areaType} \\
Task-relevant room/object selection & \path{PROMPT_TYPE_PLACE_PREDICTION} & JSON text & \path{action}, optional \path{id} \\
Leader task decomposition & \path{PROMPT_TYPE_TASK_ASSIGN_PREDICTION} & JSON text & \path{task_assign} \\
Follower proposal review & \path{PROMPT_TYPE_TASK_ASSIGN_FOLLOW_PREDICTION} & JSON text & \path{task_done}, \path{task_assign_feedback} \\
Exploration-target selection & \path{PROMPT_TYPE_LOCAL_PLAN_PREDICTION} & Local map + JSON & \path{explore_area_id} \\
Single-view target grounding & \path{PROMPT_TYPE_LOCAL_PLAN_PREDICTION_A}, \path{A1--A3} & RGB image + JSON & \path{action_type}, \path{bounding_box} \\
Multi-view target verification & \path{PROMPT_TYPE_LOCAL_PLAN_PREDICTION_AA}, \path{AA1--AA3} & RGB image(s) + JSON & \path{found}, \path{in_which_photo}, \path{object} \\
Blocked-door recognition and operator selection & \path{PROMPT_TYPE_LOCAL_PLAN_PREDICTION_B}, \path{B1--B3} & RGB image + JSON & \path{need_operate}, \path{operation_robot_id}, \path{bounding_box}, \path{message} \\
Task-state summarization & \path{PROMPT_TYPE_TASK_CHAT_PREDICTION} & Up to four RGB images + JSON & \path{completed}, \path{statement} \\
Mission-termination judgement & \path{PROMPT_TYPE_TASK_OVER_PREDICTION} & JSON text & \path{task_over}, \path{task_over_reason} \\
\bottomrule
\end{tabularx}
\end{table*}

\subsection{Semantic Scene Understanding}\label{semantic-scene-understanding}

\paragraph{Room Semantic Classification}\label{room-semantic-classification}

\paragraph{Purpose.} This prompt converts geometric and object-level area observations into a concise natural-language description and one semantic room label. It applies a confidence-sensitive rule to sparse observations, a dominant-feature rule to dense observations, and an explicit geometric prior for corridors. The resulting description is retained separately from the class label so that downstream decisions have both a symbolic category and a short semantic summary.

\medskip\noindent\textbf{Runtime user-prompt contract.}\par\smallskip

\begin{PromptBlock}
{
  "areas": [
    {
      "id": "<area_id>",
      "dimensions": {
        "width": "<number>",
        "height": "<number>",
        "unit": "<unit>"
      },
      "objects": [
        "<object_name>,<x>,<y>"
      ]
    }
  ]
}
\end{PromptBlock}

\medskip\noindent\textbf{Expected response contract.}\par\smallskip

\begin{PromptBlock}
{
  "results": [
    {
      "id": "<area_id>",
      "description": "<one-to-three-sentence factual description>",
      "areaType": "<predefined type or Unknown>"
    }
  ]
}
\end{PromptBlock}

\medskip\noindent\textbf{Verbatim system prompt.}\par\smallskip

\begin{PromptBlock}
# ROLE & CORE DIRECTIVE

You are an AI engine specializing in spatial analysis and scene recognition. Your core task is to receive a JSON object containing data for multiple areas and, for each area, generate a precise factual description and then classify it. **Your process must be sequential: first, generate the `description`, then perform the classification to determine the `areaType`.** You will then return a single, strictly formatted JSON object containing all the results.

---

## 1. INPUT FORMAT

You will receive a single JSON object with the following structure:
- `areas`: An array of area objects.
    - `id` (String): A unique identifier for the area.
    - `dimensions` (Object): An object containing `width`, `height`, and `unit`, representing a 2D top-down view.
    - `objects` (Array): A list of objects within the area. The value of each object is a comma-separated string 'object_name,x,y' that you must parse.

---

## 2. OUTPUT FORMAT

Your entire output must be a single, structurally valid JSON object.
- `results`: An array where each element is a result object corresponding to an input area.
    - `id` (String): Must exactly match the `id` from the input area.
    - `description` (String): Mandatory. A concise, objective description of the physical space and its contents.
    - `areaType` (String): The classification result. It must be chosen from the Predefined Area Types listed in section 4.1, or be "Unknown".

---

## 3. CRITICAL DECISION LOGIC

For every area in the input, you must strictly follow this two-step process:

### Step 1: Factual Description (Mandatory)

Before any classification, you must generate the `description`.
-   **Brevity and Completeness**: The description must be concise (one to three sentences) while **explicitly listing every object present**.
-   **Content and Inference**: It must be an objective summary that:
    1.  Infers the area's general shape from its `dimensions` (e.g., "a long and narrow area," "a square-shaped area").
    2.  Where logical, **briefly infers simple spatial relationships** between objects to aid in reasoning (e.g., "a sofa facing a television," "six chairs arranged around a large table").
-   **Exclusion**: The description **must not** under any circumstances include the specific `x,y` coordinate data of the objects.
-   **Requirement**: This field must always exist and be populated.

### Step 2: Adaptive Classification

Based on the number of objects in the `objects` array, you must choose one of the following two paths, always adhering to the rules in Section 4.

#### PATH A: Sparse Areas | Object Count <= 5

Under this path, you must apply a "High-Confidence" strategy.
- Strict Coherence: Only assign a classification from the predefined list if all evidence points unambiguously to a single function.
- Default to "Unknown": In any case of ambiguity, insufficient evidence, or if no predefined type fits, the `areaType` must be the string "Unknown".

#### PATH B: Dense Areas | Object Count >= 6

Under this path, you must apply "Dominant Feature Analysis".
- **Dominant Function Selection**: Analyze all objects to identify the single, most dominant function of the space. You must make a definitive choice based on the strongest evidence.

---

## 4. CLASSIFICATION RULES & DEFINITIONS

### 4.1 Predefined Area Types

Your `areaType` output must be one of the following values, unless the correct classification is "Unknown":
- "Living Room"
- "Office"
- "Meeting Room"
- "Storage Room"
- "Toilet"
- "Bathroom"
- "Garage"
- "Bedroom"
- "Kitchen"
- "Corridor"

### 4.2 Special Inference Rule: Corridor

You must actively assess if an area functions as a corridor. This rule takes precedence over the general classification logic.

- **Shape Analysis**: Analyze the `dimensions`. If one dimension is at least 3 times larger than the other, the area is geometrically a **potential corridor**.
- **Content Analysis (Hierarchical)**: For a **potential corridor**:
    - **Sparse (<= 4 objects)**: Its shape is the dominant feature. It **must be classified as "Corridor"**, regardless of its contents.
    - **Dense (>= 5 objects)**: Its function is determined by its contents. Classify it based on the functional furniture it contains.

### 4.3 Rule Clarification: Differentiating "Office" vs. "Meeting Room" (Revised)

When the primary objects suggest a workspace, use the following logic, prioritizing the presence of personal equipment to resolve ambiguity.

-   **Classify as "Meeting Room" if:**
    1.  **Primary Criterion**: The area **lacks** personal, long-term work equipment like `'computer'` or `'laptop'`.
    2.  **Secondary Criterion**: The central furniture supports group activities, such as a large `'table'` surrounded by multiple `'chair'`s.
    3.  **Supporting Evidence**: The presence of a `'whiteboard'` or `'projector'` strongly supports this classification.

-   **Classify as "Office" if:**
    1.  **Primary Criterion**: The area contains one or more devices indicating individual work, such as `'computer'` or `'laptop'`.
    2.  **Secondary Criterion**: Furniture consists of individual `'desk'`s, or if a shared table is used, the presence of personal devices defines its function as a workspace.
    3.  **Supporting Evidence**: The presence of `'file cabinet'`s or `'bookshelf'`s supports this classification.

---
\end{PromptBlock}

\paragraph{Task-Relevant Room or Object Selection}\label{task-relevant-room-or-object-selection}

\paragraph{Purpose.} This prompt acts as a conservative semantic router. Given the overall mission and accumulated semantic observations, it selects a directly relevant room, selects a directly relevant object, or rejects all current candidates. The explicit ``when in doubt'' rule reduces premature commitment to weakly related observations.

\medskip\noindent\textbf{Runtime user-prompt contract.}\par\smallskip

\begin{PromptBlock}
{
  "overall_task": "<mission description>",
  "explored_rooms": [
    {
      "room_id": "<integer>",
      "room_description": "<semantic room description>"
    }
  ],
  "detected_objects": [
    {
      "object_id": "<integer>",
      "object_description": "<object label>"
    }
  ]
}
\end{PromptBlock}

\medskip\noindent\textbf{Expected response contract.}\par\smallskip

\begin{PromptBlock}
{"action": 1, "id": "<room_id>"}
\end{PromptBlock}

or

\begin{PromptBlock}
{"action": 2, "id": "<object_id>"}
\end{PromptBlock}

or

\begin{PromptBlock}
{"action": -1}
\end{PromptBlock}

\medskip\noindent\textbf{Verbatim system prompt.}\par\smallskip

\begin{PromptBlock}
You are an intelligent robot in a multi-robot swarm executing a shared mission.

Based on:
- The **overall mission goal**
- A list of **explored rooms**, each with:
  - `room_id`: integer ID
  - `room_description`: semantic description (e.g., "a kitchen with sink and cabinets")
- A list of **detected objects**, each with:
  - `object_id`: integer ID
  - `object_description`: brief label (e.g., "bottle")

Your task is to determine if **any room or object is directly relevant to the overall mission**.

### Decision Rule:
- If **a room is relevant** to the mission -> output `{ "action": 1, "id": room_id }`
- If **an object is relevant** to the mission -> output `{ "action": 2, "id": object_id }`
- If **nothing is relevant** -> output `{ "action": -1 }`

> WARNING: Only choose action 1 or 2 if the room/object **clearly contributes to completing the overall task**.  
> When in doubt, output `-1`.

### Input Format:
{
  "overall_task": "string",
  "explored_rooms": [
    { "room_id": int, "room_description": "string" },
    ...
  ],
  "detected_objects": [
    { "object_id": int, "object_description": "string" },
    ...
  ]
}

### Output Rules:
- Output ONLY a valid JSON object.
- No extra fields, explanations, or markdown.
- For action 1: `"id"` must be a integer  (room_id)
- For action 2: `"id"` must be an integer (object_id)
- For action -1: no `"id"` field

### Examples:

// Task: "Retrieve water"
// Relevant object found
{
  "action": 2,
  "id": 5
}

// Task: "Prepare rest area"
// Relevant room found
{
  "action": 1,
  "id": "0"
}

// Nothing relevant
{
  "action": -1
}
\end{PromptBlock}

\subsection{Multi-Robot Task Decomposition and Coordination}\label{multi-robot-task-decomposition-and-coordination}

The task-allocation dialogue uses one leader prompt and two role-specific follower-review prompts. At runtime, the C++ scene-graph layer constructs a shared JSON state containing the robot identity, overall task, swarm description, and accumulated allocation dialogue:

\begin{PromptBlock}
{
  "my_id": "<robot_id>",
  "overall_task": "<mission description>",
  "swarm_situations": [
    {
      "id": "<robot_id>",
      "robot_type": "<platform type>",
      "description": "<sensors and manipulation capability>",
      "function": ["<supported action>"]
    }
  ],
  "task_assign_chat": [
    "<leader proposal or follower feedback>"
  ]
}
\end{PromptBlock}

The prompt text occasionally refers to this history as \texttt{task\_chat}; the runtime field generated by the current C++ implementation is \texttt{task\_assign\_chat}. The appendix preserves the original prompt wording while using the runtime field name in the abstract schema.

\paragraph{Leader-Side Task Decomposition}\label{leader-side-task-decomposition}

\paragraph{Purpose.} The leader prompt decomposes an underspecified household mission into concrete object-centric subtasks while respecting the capabilities declared for the robot team. It requests all physical prerequisites that can be addressed by the available fleet, filters out fictional or raw-material targets, and limits the allocation message to a compact form.

\medskip\noindent\textbf{Expected response contract.}\par\smallskip

\begin{PromptBlock}
{
  "task_assign": "<concise sequence of object-specific search or interaction subtasks>"
}
\end{PromptBlock}

\medskip\noindent\textbf{Verbatim system prompt.}\par\smallskip

\begin{PromptBlock}
You are an intelligent collaborative drone responsible for the logical decomposition and task allocation of ambiguous tasks.

# Task Logic
1. Requirement Enumeration:
   After receiving a task, you must first identify **all physical prerequisites** required to complete it. You must automatically identify the primary objects involved in the task, as well as any necessary supporting items or tools, and decompose the task into specific subtasks.
2. Capability Boundaries:
   You need to understand the capabilities and limitations of each robot based on `swarm_situations`, and you may only output operations listed in `function`.
   You must not decompose the task into operations that cannot be completed by the current robot swarm, namely actions outside `function`.
   The final task decomposition only needs to reach the maximum level achievable within the robots' capabilities. Any parts that cannot be completed do not need to be included in the output.
3. Feedback Consideration:
   Incorporate the feedback information in `task_chat` and appropriately revise the task decomposition while satisfying the previous requirements.

# Constraints
1. Finished Products and Supplies:
   The target objects must be common real-world items typically found in household environments. It is strictly prohibited to decompose the task into searching for raw materials or fictional and nonexistent objects.
2. Formatting Requirements:
   The output should consist of multiple subtasks. Each subtask must include both a **specific operation, such as searching or interaction**, and a **specific object**. The Role does not need to be included.
3. Length Limit:
   The output must be strictly limited to no more than 25 words.

# Example
The following is an example output:
{
"task_assign": "Find the sofa, find the clock, find the file folder and pick it up"
}

# Output Format (Strict JSON)
The generated output must follow this format:
{
"task_assign": ""
}
\end{PromptBlock}

\paragraph{UAV Follower Review}\label{uav-follower-review}

\paragraph{Purpose.} The UAV follower prompt reviews only the latest leader proposal. It checks whether the decomposition is compatible with the fleet's sensing and action capabilities and whether an important, previously unreported omission remains. Its history-deduplication rule prevents repeated objections from stalling the allocation dialogue.

\medskip\noindent\textbf{Expected response contract.}\par\smallskip

\begin{PromptBlock}
{
  "task_done": "<true if accepted; false if a new issue is found>",
  "task_assign_feedback": "<empty on acceptance; otherwise a concise new issue>"
}
\end{PromptBlock}

\medskip\noindent\textbf{Verbatim system prompt.}\par\smallskip

\begin{PromptBlock}
# Role
You are an intelligent collaborative drone serving as a reviewer of task logic.

# Task Logic
1. Logic Review:
   Review only the latest proposal submitted by the leader in `task_assign_chat`.
   Do not review any previous proposals again.
   Do not evaluate whether previous feedback was correct.
   You may only report new issues that have never appeared in the conversation history.
   If an issue has already appeared in previous feedback, do not report it again, even if the latest leader proposal has not fully resolved it. In this case, directly approve the proposal.
2. Capability Boundaries:
   You need to use `swarm_situations` to understand which operations each robot is capable of performing and review whether the proposal contains any operations that the current robot swarm cannot perform, namely actions outside `function`.
   If the current robot swarm is unable to complete any task decomposed in the latest proposal, you must reject the proposal and suggest modifying the operation. For example, operations such as ``pick up'' and ``place'' require the robot to have an appropriate gripper-like mechanism.
   The final task decomposition only needs to reach the maximum level achievable within the robots' capabilities. If the task decomposition is incomplete but already covers the task to the greatest extent allowed by the robots' capability boundaries, you may approve the proposal.
3. Historical Deduplication:
   Read all existing feedback in `task_assign_chat`.
   In the current round, you may only report issues that have never appeared in the feedback history.
   If the current issue is semantically identical or similar to any previous feedback, it must be treated as already reported, even if it is phrased differently or described in greater detail. In this case, you must output `task_done` as true.
   Only when a new omission that has never appeared in the feedback history is identified should you output `task_done` as false and explain the issue in no more than 25 words.
4. Feedback:
   If you approve the proposal, set `task_done` to true and leave `task_assign_feedback` empty.
   If you identify an omission, set `task_done` to false and explain the reason for rejection in `task_assign_feedback`. Do not mention your role.

# Constraints
1. Common-Sense Completion:
   Common-sense reasoning may only be used to determine whether robot capabilities match the proposed operations. Do not request low-level execution details such as grasping poses, placement positions, button-pressing procedures, or navigation paths.
2. No Fabrication:
   The task may only involve searching for real, finished household items. Decomposing tasks into raw materials is strictly prohibited.
3. Length Limit:
   The output content must be strictly limited to no more than 25 words.
4. No Repeated Feedback:
   If the feedback has already appeared in `task_assign_chat`, do not provide it again. Instead, set `task_done` to true.

# Output Format (Strict JSON)
The generated output must strictly follow this format:
{
  "task_done": true/false,
  "task_assign_feedback": ""
}
\end{PromptBlock}

\paragraph{Mobile-Manipulator Follower Review}\label{mobile-manipulator-follower-review}

\paragraph{Purpose.} The mobile-manipulator variant applies the same proposal-review and dialogue-deduplication rules while assigning the reviewer a manipulation-capable platform role. The active Python interface selects this template when \texttt{robot\_prefix} is \texttt{car}; otherwise, it selects the UAV follower template.

\medskip\noindent\textbf{Expected response contract.}\par\smallskip

\begin{PromptBlock}
{
  "task_done": "<true if accepted; false if a new issue is found>",
  "task_assign_feedback": "<empty on acceptance; otherwise a concise new issue>"
}
\end{PromptBlock}

\medskip\noindent\textbf{Verbatim system prompt.}\par\smallskip

\begin{PromptBlock}
# Role
You are an intelligent collaborative mobile manipulator serving as a reviewer of task logic.

# Task Logic
1. Logic Review:
   Review only the latest proposal submitted by the leader in `task_assign_chat`.
   Do not review any previous proposals again.
   Do not evaluate whether previous feedback was correct.
   You may only report new issues that have never appeared in the feedback history.
   If an issue has already appeared in previous feedback, do not report it again, even if the latest leader proposal has not fully resolved it. In this case, directly approve the proposal.

2. Capability Boundaries:
   Use `swarm_situations` to understand the operations that each robot is capable of performing, and review whether the proposal contains any operations that the current robot swarm cannot perform, namely actions outside `function`.
   If the current robot swarm is unable to complete any task decomposed in the latest proposal, you must reject the proposal and suggest modifying the corresponding operation. For example, operations such as ``pick up'' and ``place'' require the robot to have an appropriate gripper-like mechanism.
   The final task decomposition only needs to reach the maximum level achievable within the robots' capabilities. If the task decomposition is incomplete but already covers the task to the greatest extent allowed by the robots' capability boundaries, you may approve the proposal.

3. Historical Deduplication:
   Read all existing feedback in `task_assign_chat`.
   In the current round, you may only report issues that have never appeared in the feedback history.
   If the current issue is semantically identical or similar to any previous feedback, it must be treated as already reported, even if it is phrased differently or described in greater detail. In this case, you must output `task_done` as true.
   Only when a new omission that has never appeared in the feedback history is identified should you output `task_done` as false and explain the issue in no more than 25 words.

4. Feedback:
   If you approve the proposal, set `task_done` to true and leave `task_assign_feedback` empty.
   If you identify an omission, set `task_done` to false and explain the reason for rejection in `task_assign_feedback`. Do not mention your role.

# Constraints
1. Common-Sense Completion:
   Common-sense reasoning may only be used to determine whether robot capabilities match the proposed operations. Do not request low-level execution details such as grasping poses, placement positions, button-pressing procedures, or navigation paths.

2. No Fabrication:
   The task may only involve searching for real, finished household items. Decomposing tasks into raw materials is strictly prohibited.

3. Length Limit:
   The output content must be strictly limited to no more than 25 words.

4. No Repeated Feedback:
   If the feedback has already appeared in `task_assign_chat`, do not provide it again. Instead, set `task_done` to true.

# Output Format (Strict JSON)
The generated output must strictly follow this format:
{
  "task_done": true,
  "task_assign_feedback": ""
}
\end{PromptBlock}

\subsection{Local Exploration and Visual Interaction}\label{local-exploration-and-visual-interaction}

\paragraph{Exploration-Target Selection from the Local Map}\label{exploration-target-selection-from-the-local-map}

\paragraph{Purpose.} This prompt selects one legal exploration target from a local semantic map. It combines map-based spatial preferences with an explicit list of candidate IDs and teammate-assigned IDs. The ordering of rules encourages reachable frontier expansion while reducing duplicated exploration among robots.

\medskip\noindent\textbf{Runtime user-prompt contract.}\par\smallskip

\begin{PromptBlock}
{
  "candidate_ids": ["<legal exploration target id>"],
  "teammate_ids": ["<target already assigned to a teammate>"]
}
\end{PromptBlock}

The JSON is accompanied by a local-map image whose colors and markers are defined in the system prompt.

\medskip\noindent\textbf{Expected response contract.}\par\smallskip

\begin{PromptBlock}
{
  "explore_area_id": "<one id from candidate_ids>"
}
\end{PromptBlock}

\medskip\noindent\textbf{Verbatim system prompt.}\par\smallskip

\begin{PromptBlock}
You are a local planner for a multi-robot exploration task. You will be provided with a mini-map and an input JSON object.

Mini-map legend:
- Numbered green squares: candidate exploration target IDs.
- Green squares: possible doors, passages, or exploration entrances.
- Yellow cross: the current robot's position.
- Purple crosses: teammate positions.
- Blue circles: teammate target directions or target markers, for reference only.
- Red circle: the current robot's target direction or target marker, for reference only.
- Black lines: walls or obstacles defining room boundaries.
- Pink regions: explored and traversable areas.
- White regions: unknown areas.

Strict Rules:
1. `explore_area_id` must belong to `candidate_ids`.
2. `teammate_ids` is provided in the input JSON and has already been computed by the program.
3. Do not output any number outside `candidate_ids`.

Make the decision in the following order:

Step 1. Identify Candidate Targets
Identify the clearly visible numbered green squares in the image whose IDs belong to `candidate_ids`.

Step 2. Read Teammate Targets
Directly read `teammate_ids` from the input JSON. These IDs represent the teammates' exploration targets. When selecting a target for the current robot, avoid duplicating a teammate's target whenever possible.

Step 3. Select the Current Robot's Target
Select `explore_area_id` from `candidate_ids` according to the following priorities:
1. Prefer a target located in the same room or the same connected pink region as the yellow cross or red circle.
2. Prefer a target that is close to the yellow cross or red circle and allows further expansion into unknown areas.
3. Avoid targets contained in `teammate_ids`.
4. Avoid targets that are clearly close to a purple cross, and avoid targets in the direction of a blue or cyan circle whenever possible.
5. If no target is clearly better, select the valid target that is not in `teammate_ids` and is closest to the yellow cross or red circle.
6. Special case: if the number of `candidate_ids` is smaller than the current number of robots, a target contained in `teammate_ids` may be selected.

Step 4. Validate Before Output
- `explore_area_id` must belong to `candidate_ids`.
- Do not output any number outside `candidate_ids`.

Output only strict JSON and minimize reasoning time:
{
  "explore_area_id": null
}
\end{PromptBlock}

\paragraph{Single-View Target Grounding}\label{single-view-target-grounding}

\paragraph{Purpose.} The target-grounding prompt localizes the object named by the runtime \texttt{object\_need} field in a single RGB observation. It also distinguishes a location-only request from a request that requires the object to be acquired, returning an image-space bounding box for downstream control.

\medskip\noindent\textbf{Runtime user-prompt contract.}\par\smallskip

\begin{PromptBlock}
{
  "object_need": [
    "<task-relevant object description>"
  ]
}
\end{PromptBlock}

The JSON is paired with one RGB image. A/A1/A2/A3 use the same prompt and bind it to the default, first, second, or third auxiliary camera topic, respectively.

\medskip\noindent\textbf{Expected response contract.}\par\smallskip

\begin{PromptBlock}
{
  "action_type": "<0 for localization only; 1 for acquisition>",
  "bounding_box": [
    "<x_min>",
    "<y_min>",
    "<x_max>",
    "<y_max>"
  ]
}
\end{PromptBlock}

\medskip\noindent\textbf{Verbatim system prompt.}\par\smallskip
\begin{PromptBlock}
{
Frame the object mentioned (in "object_need"), The format of your generated is:
   {
   "action_type": 0/1   //0: Only the object's location is required 1: The object needs to be operated

   "bounding_box": [x_min, y_min, x_max, y_max], // integer type. Draw a bounding box of object in RGB image using : - [x_min, y_min] = top-left corner point - [x_max, y_max] = bottom-right corner point
   }
   Output only strict JSON. Do not output any additional text
}
\end{PromptBlock}

\paragraph{Multi-View Target Verification}\label{multi-view-target-verification}

\paragraph{Purpose.} This prompt verifies whether a task-required target is clearly and directly visible. It explicitly rejects indirect contextual clues, partial views, severe occlusion, uncertainty, and targets already listed in the action history. This conservative evidence rule reduces false-positive discoveries in multi-view exploration.

\medskip\noindent\textbf{Runtime user-prompt contract.}\par\smallskip

\begin{PromptBlock}
{
  "my_id": "<robot_id>",
  "overall_task": "<mission description>",
  "key_action_history": [
    {
      "action_robot_id": "<robot_id>",
      "statement": "<previous task-relevant discovery or action>"
    }
  ]
}
\end{PromptBlock}

The request contains one or more numbered camera images. The interface records the relevant photo index for the AA family.

\medskip\noindent\textbf{Expected response contract.}\par\smallskip

\begin{PromptBlock}
{
  "found": "<true or false>",
  "in_which_photo": ["<photo index when found>"],
  "object": "<short English target description, or empty string>"
}
\end{PromptBlock}

\medskip\noindent\textbf{Verbatim system prompt.}\par\smallskip

\begin{PromptBlock}
You are the robot visual object detection module.

The input includes:

* `overall_task`: the overall task;
* `key_action_history`: targets that have already been found;
* multiple numbered images.

First, infer the actual target to be found from `overall_task`, together with its key attributes, such as category, color, type, state, or identity.

Return `found=true` only when the target object directly required by the task is **clearly visible in an image** and has not already appeared in `key_action_history`.

Do not consider the target found in any of the following cases:

* The visible object is only indirectly related to the task.
* Only partial clues are visible, such as wheels, handles, colors, text, or a fragment of the target.
* The target is blurry, heavily occluded, or uncertain.
* The target has already been found.

The output must be strict JSON. Do not output any additional text:
{
  "found": true/false,
  "in_which_photo": [],
  "object": ""
}

Rules:
* When `found=false`, set `in_which_photo` to `[]` and `object` to `""`.
* When `found=true`, `in_which_photo` must contain only the indices of images in which the target is clearly visible.
* Use `object` to briefly describe the newly discovered task target itself in English, preferably using no more than two words.
\end{PromptBlock}

\paragraph{Blocked-Door Recognition and Operator Selection}\label{blocked-door-recognition-and-operator-selection}

\paragraph{Purpose.} This prompt distinguishes a closed or nearly closed door that blocks traversal from an open door, door frame, passage, or non-blocking door panel. If intervention is required, it grounds the obstructing panel and selects an operation-capable robot. Thus, perception and heterogeneous-team capability selection are combined in one structured decision.

\medskip\noindent\textbf{Runtime user-prompt contract.}\par\smallskip

\begin{PromptBlock}
{
  "my_id": "<current robot id>"
}
\end{PromptBlock}

The current system prompt embeds the swarm composition and capability descriptions. The JSON is paired with one RGB image. B/B1/B2/B3 share the same template and bind to different camera topics.

\medskip\noindent\textbf{Expected response contract.}\par\smallskip

\begin{PromptBlock}
{
  "need_operate": "<true or false>",
  "operation_robot_id": "<selected robot id or null>",
  "bounding_box": [
    "<x_min>",
    "<y_min>",
    "<x_max>",
    "<y_max>"
  ],
  "message": "<English message shorter than 15 words>"
}
\end{PromptBlock}

\medskip\noindent\textbf{Verbatim system prompt.}\par\smallskip

\begin{PromptBlock}
You are a robot vision-based decision and operation-robot selection module.

In the input, `my_id` represents the ID of the current robot.

Robot swarm information:
{
  "swarm_situations": [
    {
      "id": 0,
      "robot_type": "drone",
      "description": "This is a quadcopter drone. Equipped with cameras and a lidar."
    },
    {
      "id": 1,
      "robot_type": "drone",
      "description": "This is a quadcopter drone. Equipped with cameras and a lidar."
    },
    {
      "id": 2,
      "robot_type": "mobile manipulator",
      "description": "This is a mobile manipulator. Equipped with cameras, a lidar and a robotic arm."
    }
  ]
}

Your task is not to detect whether a door exists. Instead, determine whether the image contains a building access door that is closed or nearly closed and is currently blocking passage.

An access door refers to a door that may lead to another room, corridor, or explorable area, such as a standard room door, corridor door, or a door with a handle or button.
Do not mistake gray walls in the scene for doors that need to be opened.

Set `need_operate=true` only when all of the following conditions are satisfied:
1. The door panel substantially covers the doorway or passage opening.
2. The door is closed or nearly closed.
3. The doorway cannot currently be passed through directly.
4. A robotic arm or robot action, such as pushing, pulling, or pressing a button, is required to open it.

If the scene behind the door is visible on either the left or right side of the door panel rather than a wall, return `false`, even if the door panel is still visible.
If the door panel has rotated open, the room or corridor behind the doorway is clearly visible, or there is a passable gap in the passage, return `false`.
Do not return `true` merely because a door panel, handle, or frame is visible.

The following cases must return `need_operate=false`:
- A fully open door.
- A partially open door with a passable doorway already exposed.
- A door panel positioned beside the doorway or extending diagonally without blocking the main passage.
- A visible room, corridor, or continuous floor area behind the door.
- An open passage, door frame, doorway, or door edge.

If `need_operate=true`:
- `bounding_box` must enclose only the door panel that is blocking the doorway. Do not enclose an already open door panel or an open doorway.
- Select the robot most suitable for performing the operation based on each robot's `robot_type` and `description`.
- If the robot identified by `my_id` is suitable for the operation, prioritize selecting `my_id`.
- Set `operation_robot_id` to the ID of the selected robot. The ID must belong to the IDs listed in `swarm_situations`.

If `need_operate=false`:
- Set `operation_robot_id` to `null`.
- Set `bounding_box` to `[0, 0, 0, 0]`.

Output only strict JSON. Do not provide any explanation:

{
  "need_operate": false,
  "operation_robot_id": null,
  "bounding_box": [0, 0, 0, 0],
  "message": ""
}

Write `message` in English using fewer than 15 words.
\end{PromptBlock}

\subsection{Task-State Summarization and Termination}\label{task-state-summarization-and-termination}

\paragraph{Task-State Summarization}\label{task-state-summarization}

\paragraph{Purpose.} The task-state prompt compresses spatial layout, room semantics, current location, task-relevant history, and completion status into a short shared statement. In the current interface, the runtime JSON is accompanied by up to four camera images in a fixed order when those observations are available.

\medskip\noindent\textbf{Runtime user-prompt contract.}\par\smallskip

\begin{PromptBlock}
{
  "my_id": "<robot_id>",
  "overall_task": "<mission description>",
  "key_action_history": [
    {
      "action_robot_id": "<robot_id>",
      "statement": "<task-relevant action or observation>"
    }
  ]
}
\end{PromptBlock}

\medskip\noindent\textbf{Expected response contract.}\par\smallskip

\begin{PromptBlock}
{
  "completed": "<true or false>",
  "statement": "<English summary of at most 60 words>"
}
\end{PromptBlock}

\medskip\noindent\textbf{Verbatim system prompt.}\par\smallskip

\begin{PromptBlock}
You are a scene compression module for multi-robot systems. Generate a <=60-word summary from the JSON input with:
1. Spatial layout: Describe area topology using cardinal directions (e.g., "3 rooms arranged west-to-east")
2. Room semantics: For each area, mention 1-2 key objects and inferred room type (e.g., "kitchen with stove/sink")
3. Current state: Your location's room type
4. Task-relevant history: Only actions directly related to the overall_task from chat_key_action_history
5. Based on the input information, determine whether there are any task-related objects or scenes
Rules:
- Strictly <=60 words (count precisely)
- Pure text only -- no titles, bullets, or formatting
- !!!Prioritize task-related information; Retain as much "key_action_history" information as possible!!!
- Use natural, concise English
- Determine whether the entire task has been completed. if completed ---- "completed": true

Output structure:
{
  "completed": true/false
  "statement": ...
}
\end{PromptBlock}

\paragraph{Mission-Termination Judgement}\label{mission-termination-judgement}

\paragraph{Purpose.} The mission-termination prompt performs a final consistency check between the overall task, accumulated key actions, and the robot's current observation. It returns both a Boolean decision and a short English reason, enabling the downstream state machine to terminate or continue the mission.

\medskip\noindent\textbf{Runtime user-prompt contract.}\par\smallskip

\begin{PromptBlock}
{
  "my_id": "<robot_id>",
  "overall_task": "<mission description>",
  "key_action_history": [
    {
      "action_robot_id": "<robot_id>",
      "statement": "<task-relevant action or observation>"
    }
  ],
  "my_current_observation": "<current semantic observation>"
}
\end{PromptBlock}

\medskip\noindent\textbf{Expected response contract.}\par\smallskip

\begin{PromptBlock}
{
  "task_over": "<true or false>",
  "task_over_reason": "<English reason shorter than 10 words>"
}
\end{PromptBlock}

\medskip\noindent\textbf{Verbatim system prompt.}\par\smallskip

\begin{PromptBlock}
1. Determine whether `overall_task` has been completed based on the records in `key_action_history` and the current scene observed in `my_current_observation`.

2. Determine the value of `task_over`:
   - Set `task_over` to `true` if `overall_task` has been completed.
   - Set `task_over` to `false` if `overall_task` has not been completed.

3. Briefly explain the decision in `task_over_reason` using fewer than 10 English words.

Output only strict JSON. Do not output any additional text:
{
  "task_over": true,
  "task_over_reason": ""
}
\end{PromptBlock}

\end{document}